\documentclass{article} % For LaTeX2e
\usepackage{iclr2022_conference,times}

\usepackage[utf8]{inputenc} % allow utf-8 input
\usepackage[T1]{fontenc}    % use 8-bit T1 fonts
\usepackage{hyperref}       % hyperlinks
\usepackage{url}            % simple URL typesetting
\usepackage{booktabs}       % professional-quality tables
\usepackage{amsfonts}       % blackboard math symbols
\usepackage{nicefrac}       % compact symbols for 1/2, etc.
\usepackage{microtype}      % microtypography
\usepackage{xcolor}         % colors
\usepackage{standalone}
\usepackage{latexsym}
\usepackage{amsmath}
\usepackage{amssymb}
\usepackage{amsthm}
\usepackage{graphicx}
\usepackage{array}
\usepackage{tabu}
\usepackage{makecell}
\usepackage{paralist}
\usepackage{cases}
\usepackage{diagbox}
\usepackage{enumitem}
\usepackage{soul}
\usepackage{multirow}
\usepackage{verbatim}
\usepackage{tabulary}
\usepackage{booktabs}
\usepackage[mathscr]{euscript}
\usepackage{mathtools}
\usepackage{algorithm}
\usepackage{algpseudocode}
\usepackage{stmaryrd}
\usepackage{tikz-dependency}
\usepackage{subcaption}
\usetikzlibrary{automata,decorations.markings,arrows,positioning,matrix,calc,patterns,angles,quotes,calc}
\usepackage{adjustbox}
\usepackage{tabularx}
\usepackage{xspace}
\usepackage{tabulary}
\usepackage{afterpage}
\usepackage{hyperref}
\usepackage{url}
\usepackage{bm}
\usepackage{color}
\usepackage{graphicx}
\usepackage{slashbox}
\usepackage[toc,page]{appendix}
\usepackage{makecell}
\usepackage{boldline}
\usepackage{bbm}

\usepackage{wrapfig,lipsum,booktabs}

\definecolor{orange}{rgb}{1,0.5,0}
\definecolor{mdgreen}{rgb}{0.05,0.6,0.05}
\definecolor{mdblue}{rgb}{0,0,0.7}
\definecolor{dkblue}{rgb}{0,0,0.5}
\definecolor{dkgray}{rgb}{0.3,0.3,0.3}
\definecolor{slate}{rgb}{0.25,0.25,0.4}
\definecolor{gray}{rgb}{0.5,0.5,0.5}
\definecolor{ltgray}{rgb}{0.7,0.7,0.7}
\definecolor{purple}{rgb}{0.7,0,1.0}
\definecolor{lavender}{rgb}{0.65,0.55,1.0}

\definecolor{mypurple}{RGB}{111,61,121}
\definecolor{myblue}{RGB}{46,88,180}
\definecolor{myred}{RGB}{181,68,106}
\definecolor{myyellow}{RGB}{204,143,55}
\definecolor{amber}{rgb}{1.0, 0.75, 0.0}

\newcommand{\textblue}[1]{\textcolor{blue}{#1}}

\DeclareSymbolFont{extraup}{U}{zavm}{m}{n}
\DeclareMathSymbol{\vardiamond}{\mathalpha}{extraup}{87}

\newcolumntype{L}[1]{>{\raggedright\let\newline\\\arraybackslash\hspace{0pt}}m{#1}}
\newcolumntype{C}[1]{>{\centering\let\newline\\\arraybackslash\hspace{0pt}}m{#1}}
\newcolumntype{R}[1]{>{\raggedleft\let\newline\\\arraybackslash\hspace{0pt}}m{#1}}

\theoremstyle{definition}

\theoremstyle{remark}

\algrenewcommand{\algorithmiccomment}[1]{\leavevmode$\triangleright$ #1}

\setul{1pt}{.4pt}

\DeclareCaptionFont{tiny}{\tiny}

\usepackage[shortcuts]{extdash}  % for non-breaking hyphens. Stack Overflow says it's important for this to be the last package loaded.

\usepackage{blindtext}
\usepackage{graphicx}
\usepackage{capt-of}% or \usepackage{caption}
\usepackage{booktabs}
\usepackage{varwidth}
\newsavebox\tmpbox
\usepackage{amsmath,amsfonts,bm}

\def\eqref#1{equation~\ref{#1}}
\def\1{\bm{1}}

\DeclareMathAlphabet{\mathsfit}{\encodingdefault}{\sfdefault}{m}{sl}
\SetMathAlphabet{\mathsfit}{bold}{\encodingdefault}{\sfdefault}{bx}{n}

\DeclareMathOperator*{\argmax}{arg\,max}
\DeclareMathOperator*{\argmin}{arg\,min}

\newcommand{\resolved}[1]{}

\newcommand{\com}[1]{}
\newcommand{\firststep}{selective annotation\xspace}
\newcommand{\FirstStep}{Selective Annotation\xspace}
\newcommand{\Firststep}{Selective annotation\xspace}
\newcommand{\votek}{vote-$k$\xspace}
\newcommand{\Votek}{Vote-$k$\xspace}

\usepackage{xspace,mfirstuc,tabulary}

\definecolor{lightgray}{gray}{0.9}
\colorlet{soulgreen}{green!30}
\definecolor{red}{HTML}{FF0000}
\definecolor{blue}{HTML}{0000FF}
\definecolor{darkgreen}{HTML}{228B22}
\definecolor{dblue}{HTML}{007FFF}
\usepackage{pifont}

\newcommand{\cmark}{\textcolor{darkgreen}{\ding{51}}}

\usepackage{listings}

\definecolor{mymauve}{rgb}{0.58,0,0.82}
\title{
\FirstStep Makes Language Models Better Few-Shot Learners
}

\author{\textbf{Hongjin Su}$^\spadesuit$ \ \ 
        \textbf{Jungo Kasai}$^{\clubsuit\diamondsuit}$ \ \ 
        \textbf{Chen Henry Wu}$^{\heartsuit}$ \ \ 
        \textbf{Weijia Shi}$^\clubsuit$ \ \ 
        \textbf{Tianlu Wang}$^{\vardiamond}$ \ \
        \textbf{Jiayi Xin}$^{\spadesuit}$\\
        \textbf{Rui Zhang}$^{\bigstar}$\ \ 
         \textbf{Mari Ostendorf}$^{\clubsuit}$
        \ \ 
         \textbf{Luke Zettlemoyer}$^{\clubsuit\vardiamond}$
        \ \ 
        \textbf{Noah A.\ Smith}$^{\clubsuit\diamondsuit}$\ \ 
        \textbf{Tao Yu}$^{\spadesuit\clubsuit}$
       \\ 
  $^\spadesuit$The University of Hong Kong 
  \quad
  $^\clubsuit$University of Washington
  \quad
  $^\diamondsuit$Allen Institute for AI
  \\
  $^\heartsuit$Carnegie Mellon University \quad
  $^\bigstar$Penn State University 
 \quad 
  $^\vardiamond$Meta AI\\
  \\
  {\tt \{hjsu,tyu\}@cs.hku.hk,}\ \ {\tt henrychenwu@cmu.edu,} \ \ {\tt ostendor@uw.edu}\\
  {\tt \{jkasai,swj0419,lsz,nasmith\}@cs.washington.edu}
}

\iclrfinalcopy % Uncomment for camera-ready version, but NOT for submission.
\begin{document}

\maketitle

\setlength{\abovedisplayskip}{2pt}
\setlength{\belowdisplayskip}{2pt}

%\nascomment{suggest typesetting the name of the method as vote-$k$ rather than vote-k}
\vspace{-0.5cm}
\begin{abstract}
\vspace{-0.1cm}
%\tao{
%maybe highlighting the following points in abstract and introduction? 
%1) Unlike LM fine-tuning paradigm~\citep{darcy2022limitations}, selecting data to annotate is very important for few-shot in-context learning, which has been overlooked in recent studies. 
%2) Better few-shot in-context learning prompting via selecting diverse and representative examples to annotate?
%3) In-context learning with wisely-selected examples is a much better few-shot practice than other methods including fine-tuning?
%}
%\jungo{we haven't decided what name we use for the first step. Use the command, \\firststep so we can change later.}

Many recent approaches to natural language tasks are built on the remarkable abilities of large language models.
Large language models can perform in-context learning, where they learn a new task from a few task demonstrations, without any parameter updates.
This work examines the implications of in-context learning for the creation of datasets for new natural language tasks.
Departing from recent in-context learning methods, we formulate an annotation-efficient, two-step framework: \textit{\firststep} that chooses a pool of examples to annotate from \emph{unlabeled} data in advance, followed by prompt retrieval that retrieves task examples from the annotated pool at test time.
Based on this framework, we propose an unsupervised, graph-based \firststep method, \votek, to select diverse, representative examples to annotate. 
 % our pipeline decouples the two steps so that no  are necessary for any new test instance. 
%\chen{Is it the case that recent methods require additional annotations for new test instances?}
% In this paper, we formulate a two-step in-context learning framework that decouples \firststep and prompt retrieval.
% We show the importance of \firststep for in-context learning, which has been overlooked in recent work.
% Under this framework, the total annotation cost is the number of labeled retrieval targets.
Extensive experiments on 10 datasets (covering classification, commonsense reasoning, dialogue, and text/code generation) demonstrate that our \firststep method improves the task performance by a large margin. 
% (5.8\% and 5.2\% absolute gains \tao{Hongjin TODO: report relative
% improvement} on average for 18 and 100 annotation sizes respectively).
On average, \votek achieves a 12.9\%/11.4\% relative gain under an annotation budget of 18/100, as compared to randomly selecting examples to annotate.
% (13.0\% and 10.4\% relative gains on average for 18 and 100 annotation sizes respectively).
Compared to state-of-the-art supervised finetuning approaches, it yields similar performance with 10-100$\times$ less annotation cost across 10 tasks.
% \tao{too much about the two-step method. no need that many annotated data, ICL with 100 performs similar to 10k...? if use fixed 18 examples, select vs random > 10\%. Unlike LM fine-tuning paradigm, sample selection is very important for few-shot in-context learning, which has been overlooked in recent studies. the best few-shot performance, no need to annotate that many for ICL}
%Our extensive experiments on diverse 10 datasets demonstrate that in-context learning outperforms state-of-the-art finetuning methods by \jungo{Hongjin's TODO} on average, under the same annotation budget.
%Further, in-context learning achieve similar performance to the finetuning models with 5-8x less annotation cost. 
%\rui{These two sentences are not very clear: 1st: "ICL outperforms SOTA finetuning" vs 2nd:"ICL similar to finetuning with less annotation cost". We can merge them to deliver a clear and strong message. It is also good to have concrete numbers, e.g., resulting in up to 5-8x reduction in annotation cost without performance loss.}
%Through this analysis, we find that a small number of \emph{carefully-selected} examples suffice to perform (e.g., 100 training examples) on par with the full training data, across 10 diverse tasks in natural language processing.
%Extensive analyses show that our method improves in-context learning across language models with varying sizes and the improvement is even more pronounced when there is a domain shift in the evaluation data.
We further analyze the effectiveness of our framework in various scenarios: language models with varying sizes, alternative \firststep methods, and cases where there is a test data domain shift.
We hope that our studies will serve as a basis for data annotations as large language models are increasingly applied to new tasks.\footnote{Our code is available at \url{https://github.com/HKUNLP/icl-selective-annotation}}
%\tao{a better phrase for \firststep? \rui{Is "coreset selection" better?}}
%\weijia{People mainly use ICL for few-shot learning and may care more about few-shot setting. Maybe we could emphasize that even in true few-shot setting without prompt retrieval (training data pool size = number of ICL examples), our method is also very effective. }
% \footnote{Our code is available at \url{}.}

\begin{figure}[h!]
\vspace{3mm}
\centering
    \includegraphics[width=0.98\textwidth]{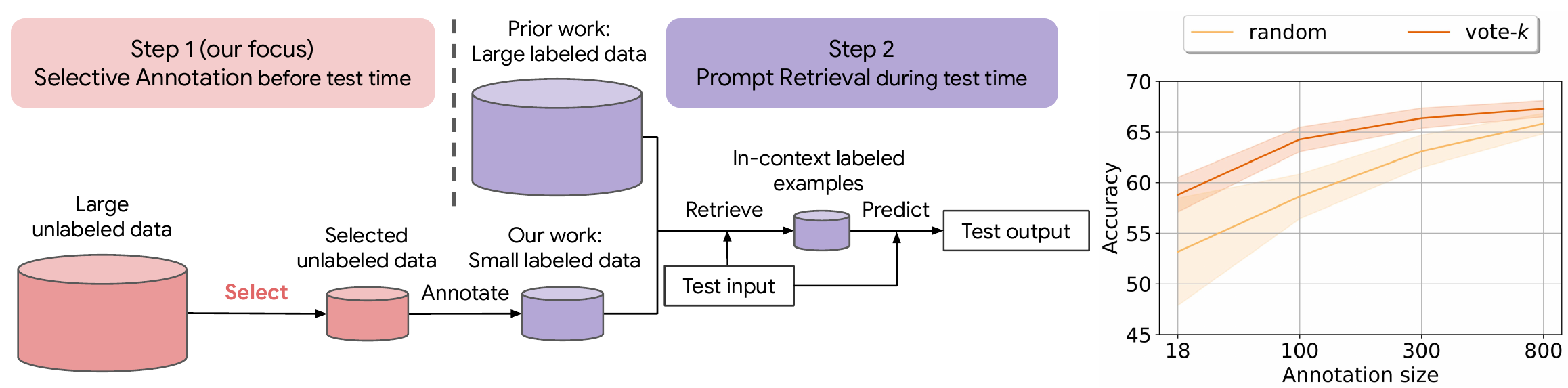}
\caption{
\textbf{Left}: Our two-step framework for in-context learning. 
Instead of assuming access to large labeled data, we first select a small number of (diverse and representative) unlabeled examples to annotate before test time.
At test time, we retrieve in-context examples from the small annotated pool.
% We show the importance of \emph{\firststep} for in-context learning.
%, which has been overlooked in recent work.
\textbf{Right}: In-context learning performance over varying annotation budgets averaged over three representative tasks (HellaSwag commonsense reasoning, MRPC paraphrase detection, and MWOZ dialogue state tracking).
%; results on 10 datasets in Tab.~\ref{tab:main_results}).
Here we experiment with GPT-J and Codex-davinci-002.
%(results on other models in Tab.~\ref{fig:model_size}). 
Two \firststep methods are presented: \textit{random selection} and our \textit{\votek} method.
%(\S\ref{sec:sample_selection_method}).
% Detailed experimental details will be discussed in \S\ref{sec:in-context_finetuning}.
%Results demonstrate that
We observe that an appropriate \firststep method largely improves the in-context learning performance with smaller variance over random selection under varying annotation budgets.
%\jungo{nitpick: can you change select to \votek for consistency?}
% In-context learning with wisely-selected labeled samples is a much better few-shot practice than other methods.
%\tao{Chen and Jungo TODO: maybe too long? edit to be more concise?} 
%\chen{I edited some text. Still very long, but maybe okay?}
%\chen{figure: vote-k -> \votek}
}
\label{fig:pipeline-avg-main-results}
\end{figure}

\end{abstract}

\section{Introduction}
Much recent work builds approaches to natural language tasks on the impressive abilities of large language models (e.g., GPT-3; \citealp{gpt3}).
Large language models can perform downstream tasks by conditioning generation on a few task demonstrations, thereby avoiding the need for any parameter updates.
This new, few-shot learning paradigm is called \textit{in-context learning} and has become an attractive alternative to supervised finetuning \citep{prompt_survey}. 
In this work, we study the implications of this remarkable capability of large language models for dataset creation and annotation.
We extensively examine how to reduce the manual annotation cost while retaining high in-context learning performance.

%A flurry of recent work \interalia{liu-etal-2022-makes,hu_dialogue} proposed methods for in-context learning that can achieve competitive performance to finetuning in many tasks, including question answering \citep{berant-etal-2013-semantic,joshi-etal-2017-triviaqa,kwiatkowski2019natural}, sentiment analysis \citep{socher-etal-2013-recursive}, and semantic parsing~\citep{Rajkumar2022EvaluatingTT}.
%It thus requires no updates to the model parameters.
% In-context learning is an attractive alternative to supervised finetuning, as it allows the large language model to avoid expensive finetuning and to adapt to new scenarios with a few in-context examples for prompting \citep{prompt_survey}. 
%, and dialogue state tracking \citep{budzianowski-etal-2018-multiwoz}.
%
Although in-context learning was originally proposed for few-shot learning, recent works show that retrieving prompts from a large set of annotated examples is necessary to achieve good performances \citep{liu-etal-2022-makes,rubin2022}. 
In particular, they show that the performance substantially improves when similar examples (under some embedding function) are retrieved as in-context examples specifically for each test input \citep{liu-etal-2022-makes}.
%diverging from the original few-shot scheme in terms of the required annotation cost \citep{gpt3}.
% It is not clear which examples should be selected to annotate to make in-context learning work for new tasks that a large annotated dataset is not available.
% Even for tasks with a large labeled training data, how many labeled examples are really needed to achieve decent in-context learning performance has been also understudied. 
Each test sample only requires a few in-context examples in its prompt.
Different test instances, however, require different in-context examples with their associated annotations, necessitating a large set of annotated examples.

%In this work, we present in-depth studies on the effect of \emph{\firststep} for in-context learning.
%We investigate how to select examples to annotate to make in-context learning work better on new tasks for which a large annotated dataset is not available.
%Even for tasks with large labeled training sets, we study how many labeled examples are really needed to achieve strong in-context learning performance. 
Distinct from these recent efforts, we establish a two-step framework to better understand and improve the annotation efficiency (Fig.\ \ref{fig:pipeline-avg-main-results}): the first step is \textit{\firststep} that picks a small number of instances to get annotated before test time, followed by \textit{prompt retrieval} that retrieves in-context examples for each test instance from the annotated data.
The total annotation budget is the number of examples selected and annotated in the first step.
The second step is bounded by the number of examples that can fit as input to a language model. Based on this framework, we propose an unsupervised, graph-based \firststep method, named vote-$k$, that selects diverse and representative instances to be annotated. 
% the number of retrieval targets corresponds to the annotation budget.
% Under this framework, prior methods that select in-context examples from all training data (e.g., \citealp{liu-etal-2022-makes,hu_dialogue}) can be viewed as a special case where the annotation budget is the same as the size of the original training data, which is generally very large (e.g., 10K). 
% \tao{more our findings and result analysis instead of the framework}

% Our framework \tao{check} enables us to systematically compare the annotation cost for in-context learning methods.
%We find that, as shown in Fig.\ \ref{fig:pipeline-avg-main-results}, \firststep largely improves in-context learning over randomly-selected annotation.
%\citep{darcy2022limitations}. 
% To reduce the annotation cost, we propose an \firststep method, \votek, that selects a diverse and representative candidate set to annotate, based on the graph structure of unlabeled samples (\S\ref{sec:sample_selection_method}).
Our extensive experiments over 10 datasets across diverse tasks (covering classification, commonsense reasoning, dialogue, and text/code generation; see Tab.~\ref{tab:main_results}) demonstrate that our graph-based \firststep method, \votek (\S\ref{sec:sample_selection_method}), substantially improves the in-context learning performance by balancing the diversity and representativeness of annotated samples. % (\S\ref{sec:div_rep_analysis}).
For instance, \votek, combined with similarity-based prompt retrieval \citep{liu-etal-2022-makes,rubin2022}, 
achieves a 11.4\% relative gain under a budget of 100 annotated examples and a 12.9\% relative gain when only 18 examples are annotated; 18 samples can fit into language models' input, meaning the prompt retrieval step is not needed.
Moreover, the improvement is consistent across language models with varying sizes (2B-175B parameters) (\S\ref{sec:lm_sizes}).
This finding is in contrast with finetuning, where we cannot see the effectiveness of \firststep over random baseline, due to outliers~\citep{karamcheti-etal-2021-mind} or training instability \citep{darcy2022limitations}.
%(Fig.~\ref{fig:icl-vs-ft}).
%and the finetuning instability over limited data~\citep{darcy2022limitations} (Fig.~\ref{fig:icl-vs-ft}).
We hypothesize that in-context learning \textit{with similarity-based prompt retrieval} is more robust to small annotation sizes and outliers because only the most similar examples are retrieved for each test instance.
Indeed, we observe that \textit{random prompt retrieval} fails to benefit from \firststep (\S\ref{sec:random_retrieval}), providing support for our hypothesis.

% 7.8 accuracy points on the SST-5 sentiment analysis task \citep{socher-etal-2013-recursive} and 9.9 points on the GeoQuery semantic parsing task \citep{geoquery96}.

%Furthermore, we show that wisely-selected annotated examples make in-context learning better and robust.
Besides performance comparisons within a fixed annotation budget, we show that \firststep provides better few-shot performance with 5-100$\times$ \textit{less annotation cost} for new natural language tasks.
%As shown in Tab.~\ref{tab:main_results}, 
In-context learning with 18 examples selected by \votek achieves higher performance than 100 randomly selected examples on 6 out of the 10 tasks.
%5-10x less annotation cost.
%\jungo{delete with...? Unpack this sentence?}
It also outperforms strong finetuning methods by a large margin (Fig.\ \ref{fig:icl-vs-ft}) 
% \chen{check: are we using sota few-shot finetuning methods or just plain finetuning?} 
and requires 10-100$\times$ less annotations for similar performance (\S\ref{sec:in-context_finetuning}).
%Moreover, Fig.\ \ref{fig:main_results} indicates that
We observe that in-context learning quickly (100 or 300 samples are annotated) converges to decent performance when \votek \firststep is applied.
These results suggest that large language models do not require large annotated datasets (e.g., 10K) due to their ability to adapt to new tasks through simple prompting.
%\tao{check}
% \tao{it would be nice to show GPT-J/XNeo (5x smaller model size) with wisely selected examples outperforms GPT-3 with random annotated examples! or as Noah: we match the performance of [something amazing] with only 1--20\% as much annotated data, across [large number] tasks}

%\jungo{Add: Since in-context learning is very unstable depending on the selected examples, standard deviations might be important to show. And indeed, the stability of our \votek method is one of its strengths. We need to add this discussion to the paper (esp. intro).}
\Firststep also makes in-context learning much more \textit{stable}. 
%Notably
%Given a set of unlabeled data, our \votek \firststep method is deterministic, without introducing randomness.
In real-world scenarios, even collecting \textit{unlabeled} data is non-trivial and introduces randomness.
We simulate such randomness in our experimental setting by subsampling the original unlabeled data multiple times.
Our results suggest that \votek \firststep largely reduces the variance of in-context learning even in this setting 
(Tab.~\ref{tab:main_results}).
% (Tabs.~\ref{tab:main_results} and \ref{all_sample_selection_method}). 
% (Appendix \ref{app:main-results}).
%Not only can our framework achieve consistent improvement over language models with varying sizes (2-175B parameters),%\chen{I moved this observation to an early paragraph. } 
Further analysis shows larger improvements when there is a domain shift between training and test data (e.g., text from different Amazon users; \citealp{wilds}; \S\ref{sec:domain_shift}).
%Our analysis illustrates that \votek selection is able to find a balanced middle ground between diversity and representativeness, which outperforms all other annotation or active learning methods in the in-context learning paradigm.
Finally, when compared with previous \firststep methods designed for supervised training/finetuning, we demonstrate that \votek \firststep consistently improves the performance 
% and reduces its variance 
(\S\ref{subsec:sample-selection-methods}).
As in-context learning has been applied to increasingly more natural language processing applications, we hope that our annotation-efficient framework will provide useful guidance for both researchers and practitioners.
% \rui{The last three paragraphs in intro contains a lot of messages which is a bit unorganized, e.g., we jump back and forth from Fig 1 and Tab 1, making it hard for readers parse. How about we first list several key findings in plain English, and then present evidence for each?}
\section{\FirstStep for In-Context Learning}
\label{sec:framework}

% \rui{For section title, is it better to use "Selective Annotation for In-Context Learning"? Basically, our job is to highlight selective annotation but still explains our two-step framework?}
% \begin{figure}[h!]
% \centering
%     \includegraphics[width=0.45\textwidth]{images/in-context_framework.pdf}
% \caption{Our two-step framework for in-context learning. Instead of assuming the access to a large labeled training data, we first select a small number of examples to annotate, and then retrieve input-specific in-context examples from the annotated data pool. \tao{to update the figure} \luke{Move this to be Fig 1 and put at beginning (where already referenced in intro)? Really helpful to see flow all in one place.}}
% \label{fig:framework}
% \end{figure}
% In-context learning uses a pretrained language model to generate a prediction for every test instance, conditioned on a few annotated, demonstration (i.e., in-context) examples \emph{without} any parameter updates \citep{gpt3}.

In-context learning  only requires a few annotated examples per test instance (\emph{few-shot learning}), while avoiding expensive finetuning on the whole training data.
It is, however, often assumed that all \emph{annotated} training data are available for prompt retrieval (e.g., \citealp{liu-etal-2022-makes,rubin2022}).
Yet the implied total annotation costs are hardly discussed in previous work.
We develop a better practice for few-shot learning with large language models by carefully studying the total annotation cost required for in-context learning.
We also study how examples should be selected to annotate, in order to make in-context learning perform better for new tasks.
We formulate a general framework (Fig.\ \ref{fig:pipeline-avg-main-results} left) that consists of two steps: \firststep (\S\ref{sec:sample_selection}) and prompt retrieval (\S\ref{sec:prompt_retrieval}).
% \tao{use math notations?}

%\subsection{General Framework}
% The original in-context learning scheme uses as a prompt a fixed set of a few randomly-sampled demonstration examples \citep{gpt3}.
% Subsequent work proposed methods to improve the performance by retrieving examples for each test instance \citep{liu-etal-2022-makes,rubin2022} at the expense of the increased annotation cost, as quantified under our framework (Fig.\ \ref{fig:framework}).

\subsection{\FirstStep}
\label{sec:sample_selection}
The first step chooses examples to annotate \emph{before} test time.
This process thus determines the total annotation budget.
% Under the original setting of in-context learning \citep{gpt3}, the \firststep size is equal to the number of randomly-sampled examples used for all test instances (typically 4-32 examples).
% In contrast, later improvements that retrieve similar examples per test instance from the full training data implicitly assume all training data are accessible.
This \firststep process is largely ignored in the recent literature for in-context learning.
We will demonstrate, however, that the annotation cost can be substantially reduced by choosing a small set of diverse, representative examples, while retaining the downstream performance (\S\ref{sec:experiments}). Formally, given a set of unlabeled samples $\mathcal{X} = \{x_i\}_{i=1}^{N}$, \firststep aims at selecting a subset $\mathcal{L} \subset \mathcal{X}$ to be annotated, where $|\mathcal{L}| = M$ is the annotation budget. 
%\tao{Chen and Jungo TODO: fix notation! e.g., $V$ here is different from $V$ below...}
We discuss our \votek \firststep method and other \firststep baselines below.
% \subsection{\Votek \FirstStep Method}
% % \rui{I still feel it's better to move this subsection in Section 2.}\luke{+1}
% At the core of in-context learning are the two steps of \firststep and prompt retrieval, as formulated in \S\ref{sec:framework}.
% %We provide three and two variants for these two steps, resulting in a total of six configurations. 
% Our experiments will show that the \firststep method is important to reduce the annotation cost, even though this step is largely ignored by recent work on in-context learning.
%\citep{liu-etal-2022-makes,rubin2022}.

\paragraph{\Votek}
\label{sec:sample_selection_method}
%Here we introduce our \votek algorithm for \firstep. 
%Here we discuss three main methods for sample selection that we use across the 10 tasks. 
%\tianlu{Not sure if it is a good way to mix these three methods. Usually I tend to look for the proposed method in the Method section and the baselines or competing methods in the Experiment section. }
%\weijia{+1 to Tianlu. It may be a little more clear to introduce the proposed method in a different section}
%We will explore more methods from the active learning literature in our analysis \S\ref{}.

%\paragraph{\Votek}
%One potential problem of the embedding diversity method is that it lacks \textit{representativeness}: a diverse set of examples might overrepresent rare outliers from the unlabeled data, and these outliers will differ from most of the test instances.
%Indeed, previous work found that many \emph{active learning} methods tend to choose and annotate outliers, resulting in performance degradation on test data \citep{karamcheti-etal-2021-mind}.
%While this work focuses on in-context learning,
%We will find that representativeness is important for in-context learning as well.
% \tao{also mention how selection goals for ICL are different from others? use this as the motivation for developing our method: diversity and representatives. not talking about the method directly without why}
The goal of \firststep for in-context learning is to select diverse and representative examples; %which after being retrieved during prompt retrieval, serve as better demonstrations for in-context learning. 
representativeness will help many test instances to find similar demonstrations, while diversity increases the total coverage.
%\jungo{I wrote the preceding sentence to give the reader some intuition. It might be a bit confusing. Is there a better way to say this?}
%examples to annotate so that it is more likely to retrieve similar in-context examples for each test instance \tao{check}.
% This is different from uncertainty-based active learning and coreset selection methods for the finetuning paradigm.
% \jungo{delete this preceding sentence? It's not clear how it's different.}
We develop \votek, a graph-based method that promotes both diversity and representativeness.
A detailed algorithm can be found in Appendix~\ref{sec:details-sample-selection}. 
We first compute a vector representation for each \emph{unlabeled} training instance using Sentence-BERT \citep{reimers-gurevych-2019-sentence} by averaging the resulting vectors over the text input words.\footnote{\url{https://huggingface.co/sentence-transformers/all-mpnet-base-v2}.}
%based on the well-established PageRank algorithm \citep{Page1999ThePC}.
We then use the embedding vectors to create a directed 
%\jungo{I think this is undirected. Is this a typo?} 
%\hongjin{The edge is directed because it is from a vertex to its k nerest neighbor}
graph $G = (V, E)$ where the vertices $V$ are the unlabeled instances $\mathcal{X}$ 
%\nascomment{note notation change; use script for datasets, switch to $V$ only when we talk about vertices}
as defined above.
For each vertex $v \in V$, we create an edge to its $k$ nearest vertices in terms of the cosine similarity between the embeddings.
 Now let $\mathcal{L}$ and $\mathcal{U}$ denote the sets of already chosen (i.e., labeled) samples and remaining samples, respectively. 
 Initially, $\mathcal{L}=\emptyset$.
 Every vertex $u \in \mathcal{U}$ is scored by a modified degree: 
 \begin{align*}
 \mathrm{score}(u) = \sum_{v \in \{v | (v, u) \in E, v \in \mathcal{U}\}} s (v), \quad \text{where} \ s(v) = \rho ^{- |\{\ell \in \mathcal{L}| (v, \ell) \in E \}|}, \quad \rho > 1
 \end{align*}
where $s$ discounts $v$ that is close to the already selected instances, thereby encouraging diversity.
In every iteration, we take $\argmax_{u \in \mathcal{U}} \mathrm{score}(u)$ and move it from $\mathcal{U}$ to $\mathcal{L}$.
We run $M/10$ of these iterations; after this process, the current labeled $\mathcal{L}$ has $M/10$ samples (up to Line~\ref{line:first} in Algorithm~\ref{alg:vote-k}).
Subsequently, we use $\mathcal{L}$ as the in-context learning examples for large language model, e.g.,GPT-J \citep{gpt-j}, and generate a prediction for every instance in $\mathcal{U}$.
We then compute the average log probability over the generation output as the model's confidence score (Line~\ref{line:lm-start} to Line~\ref{line:lm-ends} in Algorithm~\ref{alg:vote-k}).
%\nascomment{this is somewhat unclear and doesn't match what's in the algorithm in the appendix.  when you say ``average log probabilities over the output tokens'' I am unclear about what the input tokens are ... I think you need to walk through what gets fed into the LM and what scores are derived from it}
%for all instances in $\mathcal{U}$.
We then partition $\mathcal{U}$ into $M$ equal-sized buckets, based on their confidence scores (e.g., if $M=100$, we group the unlabeled instances by percentile).
We add to $\mathcal{L}$ the example with the maximum score from each of the first $9M/10$ buckets (discarding the $M/10$ buckets with the most confident examples), resulting in $|\mathcal{L}|=M$ (Line~\ref{line:it-start} to Line~\ref{line:it-ends} in Algorithm~\ref{alg:vote-k}). This further encourages diversity by selecting instances with varying confidence scores from in-context learning.
We tuned $k$ and $\rho$ in our preliminary experiments, and found that $k\!=\!150$ and $\rho\!=\!10$ perform well across many datasets.
We will explore other \firststep methods from prior work on active learning or coreset selection (\S\ref{subsec:sample-selection-methods}) and see that \votek outperforms these alternative methods.

\paragraph{Random and Other \FirstStep Methods}
To quantify the effect of \firststep, we also provide random and other baselines.
For randomly-selected annotation, we conduct experiments three times and report the average score.
We will show that these baselines substantially underperform the \votek method (\S\ref{sec:results}), demonstrating the importance of the \firststep step to reduce the total annotation cost.

\subsection{Prompt Retrieval}
\label{sec:prompt_retrieval}
Once we have a set of annotated examples $\mathcal{L}$ from \firststep, we retrieve a few examples from the annotated set as in-context examples for each test instance.
%The original in-context learning method used a fixed set of in-context examples for all test instances, meaning that prompt retrieval is not performed. 
%\tao{maybe shorten a bit, or remove this sentence?}
% Recent work \citep{liu-etal-2022-makes,rubin2022} improved the performance by retrieving similar examples to each test instance in some embedding space (e.g., BM25, \citealp{bm25}; RoBERTa, \citealp{Liu2019RoBERTaAR}; Sentence-BERT, \citealp{reimers-gurevych-2019-sentence}).
Following recent work~\citep{liu-etal-2022-makes}, we will compute embeddings for all annotated samples using Sentence-BERT and find the most similar examples to each test instance in terms of cosine similarity.
%Note that our retrieval step is also compatible with sentence embedding models finetuned on specific domains \citep{rubin2022}, but we use Sentence-BERT for simplicity and generality. 
%\chen{todo: Sewon suggested citing \citet{liu-etal-2022-makes} only for the Sentence-BERT practice, which makes a lot of sense. I wrote the last sentence in order to cite \citet{rubin2022}, but I'm not sure whether it is suitable (it seems to emphasize our limitation of the choice of prompt retrievers). }
% \chen{todo: check citation as Sewon suggested. }
%Once a small number of instances are annotated, we perform in-context learning on every test instance by retrieving examples from the pool of annotated instances.
%Previous work showed that in-context examples that are similar to the test instance generally improve the performance. We use an embedding-based method for prompt retrieval. 

%\paragraph{Embedding Similarity}
%Similar to \firststep, we apply Sentence-BERT to every example and use the cosine similarity.
%\hongjin{since test label is not accessible, I also use training instance input to calculate embeddings, so the encoding texts from training and test example are in the same format}
% The one difference is each example comes with its annotation.
% We therefore create a prompt that includes each example and its annotation and encode the entire prompt instead.
% For example, we encode the following for the first example from SST-5 in Table \ref{tab:datasets}: 
% \textit{How do you feel about the following sentence? A very well-made, funny and entertaining picture. Answer: Very Positive}.

\section{Experiments}\label{sec:experiments}
% Our framework (\S\ref{sec:framework}) enables us to systematically compare the total annotation cost for different in-context learning methods.
We conduct extensive experiments 
% with 4 in-context learning models \tao{check} %and 6 \firststep methods 
over 10 diverse datasets, spanning 9 distinct tasks, and show a better approach to few-shot learning than previously considered.
% \chen{In this section, we only have results for 2 \firststep methods?}
In general, we find that the first step of \firststep is particularly crucial to reduce the amount of required annotation.

\subsection{Datasets and Tasks}
\label{sec:datasets_tasks}
% \rui{Is it better to have a table listing: dataset name, task, task formulations(classification, multiple choices, ...), used models (GPT-3, ...). I feel this is a better presentation for information in 3.1-3.2. Or we can augment Table 1 with their corresponding ICL models?}
We use 10 diverse NLP datasets across 9 tasks that are listed in Table~\ref{tab:datasets_model}.
% commonsense reasoning \textbf{HellaSwag} \citep{zellers-etal-2019-hellaswag}, paraphrase detection \textbf{MRPC} \citep{mrpc}, sentiment analysis \textbf{SST-5} \citep{socher-etal-2013-recursive}, dialogue state tracking \textbf{MWoZ 2.4} \citep{budzianowski-etal-2018-multiwoz}, semantic parsing \textbf{GeoQuery} \citep{geoquery96}, topic classification \textbf{DBpedia} \citep{dbpedia15}, natural language inference \textbf{MNLI} \citep{williams-etal-2018-broad} and \textbf{RTE} \citep{rte1,rte2,rte3,rte5}, open-domain question answering \textbf{Naural Questions} (\textbf{NQ}; \citealp{kwiatkowski2019natural}), and summarization \textbf{XSUM} \citep{xsum2018}.
These datasets involve different task formulations, thereby allowing for extensive evaluations in varying scenarios.  % classification (MRPC, SST, DBpedia, MNLI, and RTE), multiple-choice selection (HellaSwag), dialogue state tracking (MWoZ), and code/text generation (GeoQuery, NQ, and XSUM).
Some of those are included in the widely-used GLUE benchmark \citep{wang2019glue}.
Appendix \ref{sec:datasets_all} illustrates details of the 10 datasets with examples.

For each dataset, we use the standard train/dev./test split available from the Transformers library \citep{wolf-etal-2020-transformers}. In the \firststep step, we remove all labels in the training data. 
For the datasets that have test data available publicly, we use the the test data for evaluation (SST-5, XSUM, MWoZ, and DBpedia).
For the others, we follow prior work (e.g., \citealp{jiang-etal-2020-smart,Lan2020ALBERT,gao-etal-2021-making}) and use the dev.\ data for evaluation.\footnote{The one exception is GeoQuery, where we concatenated the dev.\ and test data to have reliable evaluations on larger data.}
We evaluate the methods by accuracy for all classification and multiple-choice selection datasets, joint accuracy \citep{budzianowski-etal-2018-multiwoz} for MWoZ, test suite accuracy \citep{zhong-etal-2020-semantic} for GeoQuery, exact matching \citep{rajpurkar-etal-2016-squad} for NQ, and ROUGE-L \citep{Lin2004ROUGEAP} for XSum.
% Test suite accuracy is a well-establish metric that measures the semantic accuracy of SQL queries over a suite of high-quality text-to-SQL pairs and reduces false negatives in evaluation \citep{zhong-etal-2020-semantic}.

\begin{table*}[!h]
\begin{adjustbox}{width=0.96\linewidth}
\begin{tabular}{@{}l@{}l@{}c@{}c@{}}
\toprule
& Dataset & Task & In-Context Learning Models \\ 
\midrule[.005em]
\multirow{5}*{\textbf{Classification}} & MRPC \citep{mrpc} & Paraphrase Detection
&
GPT-Neo, GPT-J, GPT-3
\\
& SST-5 \citep{socher-etal-2013-recursive} & Sentiment Analysis & GPT-J
\\
& DBpedia \citep{dbpedia15} & Topic Classification & 
GPT-J
\\
& MNLI \citep{williams-etal-2018-broad} & Natural Language Inference & 
GPT-J
\\
& RTE \citep{rte5} & \ \ Natural Language Inference \ \ & 
GPT-J
\\
\midrule[.005em]
\multirow{1}*{\textbf{Multiple-Choice}} \ \  & HellaSwag \citep{zellers-etal-2019-hellaswag} & Commonsense \par Reasoning &
OPT, GPT-Neo, GPT-J, GPT-3
\\
\midrule[.005em]
\multirow{1}*{\textbf{Dialogue}} & MWoZ 2.4 \citep{budzianowski-etal-2018-multiwoz} & Dialogue State Tracking & 
Codex-\{cushman, davinci-002\}
\\
\midrule[.005em]
\multirow{3}*{\textbf{Generation}} & GeoQuery \citep{geoquery96} & Semantic Parsing &
Codex-davinci-002
\\
& NQ \citep{kwiatkowski2019natural} & Open-Domain QA & 
Codex-davinci-002
\\
& XSUM \citep{xsum2018} & Summarization & 
GPT-J
\\
\bottomrule
\end{tabular}
\end{adjustbox}
\caption{
All the 10 datasets and the in-context learning models used in our experiments. GPT-J and Codex-davinci-002 are used by default.
Other in-context learning models are explored in analysis. 
% \chen{todo: add OPT for some tasks. } %\chen{todo: optimize table position. }
}
\label{tab:datasets_model}
\end{table*}

% \rui{I don't know this before, but I feel this paragraph should be treated with cautions, and it should be moved to somewhere else.}
%\paragraph{Subsampling for unlabeled data}
\paragraph{Measuring Stability}
Given a set of unlabeled data, our \votek \firststep algorithm is \emph{deterministic}, without any randomness.
However, we note that in real scenarios, even getting \textit{unlabeled} samples is not trivial, and getting unlabeled samples can be a process with large variance. To simulate this real setting, we perform \firststep from 3K instances that are randomly subsampled from the original training data for each task.
% Our analysis in \S\ref{sec:in-context_finetuning} also verifies that retrieving from full training data only has marginal gain over subsampling when our \votek is used, indicating that in-context learning is annotation-efficient under our framework. \chen{Check: this is based on the file ``images/with\_full.pdf''; not sure where the reference would be finally. } \chen{Removed. I misunderstood that figure. }
% \rui{Motivation unclear: How does this save computational cost?}
% Notice that the improvement of \firststep would be even higher if we select the whole unlabeled training data.
%\rui{Can you provide experiments or some explanations to back up this claim?}
For each experiment, we repeat this subsampling three times, and results are averaged over the three trials. 
We will find that \votek still substantially improves stability over alternative \firststep methods.
%\tao{Chen and Hongjin TODO: add more details. agree with Rui}.
%\rui{I am a bit confused by this. Does this mean that \votek should also have some variance error bars in Figure 1?}

\subsection{In-Context Learning Models}
We mainly perform experiments using GPT-J with 6B parameters \citep{gpt-j} due to our computational budget.
The exceptions are the MWoZ, GeoQuery, and NQ datasets, where we use Codex-davinci-002 \citep{codex},\footnote{The parameter size of Codex is not officially confirmed, but it is likely to be 175B.} a variant of GPT-3 finetuned on code data from the web.
Codex is particularly effective for structured prediction such as semantic parsing, and we found it is indeed effective on three datasets (MWoZ, GeoQuery, and NQ) in our preliminary experiments.
We will explore the effectiveness of selective annotation on the largest publically available language models, OPT-175B~\citep{Zhang2022OPTOP} for HellaSwag (Fig.~\ref{fig:hellaswag_opt}) and Codex-davinci-002 for MWoZ, over varying annotation budgets.
We will also explore other language models with different sizes for three representative tasks (HellaSwag, MWoZ, and SST-5) in \S\ref{sec:lm_sizes}: GPT-3 with 175B \citep{gpt3} and GPT-Neo with 2.7B parameters \citep{gpt-neo}.
%\tao{adding GPT-XNeo 30B}
Our later experiments will show the same patterns among \firststep methods over these different language models.
For the classification and multiple-choice tasks, we compute the average log score for each choice and choose the maximum one.
For generation tasks, we simply perform beam-search decoding.

% To prompt these models, we create prompt templates for the 10 datasets, based on previous work \citep{bach2022promptsource}.
See Appendix \ref{sec:prompt_templates} for our in-context learning prompt templates for all 10 datasets.
For every test instance, we feed as much retrieved samples as possible into the language model until the maximum token length is reached. 
%For every test instance, we find the top-$N$ similar examples and use them as in-context examples, where $N$ is the largest number until the maximum token length of the language model is reached.
On average, the number of samples $N$ fed into the language model is 13.4 across different experiments.
The in-context examples are concatenated in the ascending order of the similarity so that more similar examples benefit from the recency bias~\citep{lu-etal-2022-fantastically}. 

\subsection{Main Results}
\label{sec:results}

\begin{table}[h!]
\addtolength{\tabcolsep}{-4.7pt}
\centering
\small
\begin{tabular}{@{} cc   m{0.001em}  ccccc   m{-0.05mm}   c m{-0.05em}  c m{0.001em}   ccc @{}}
\toprule[.1em]

 \multicolumn{2}{c}{\textbf{Method}}
&& \multicolumn{5}{c}{\textbf{Classification}}
&& \multicolumn{1}{c}{\textbf{Multi-Choice}}
&& \multicolumn{1}{c}{\textbf{Dialogue}}
&& \multicolumn{3}{c}{\textbf{Generation}}
\\
\cmidrule(lr){1-2}
\cmidrule(lr){4-8}
\cmidrule(lr){9-10}
\cmidrule(lr){12-13}
\cmidrule(lr){14-16}

$|\mathcal{L}|$
&Selection
% & Retrieval
&& MRPC & SST-5 & MNLI & DBpedia & RTE
&& HSwag
&& MWoZ 
&& GeoQ & NQ & XSum
\\

\midrule[.1em]

%Random
%& Random
%&& 62.3
%& 37.6
%& 36.2
%&  79.2
%&  55.2
%&& 63.2
%&& 43.8 
%& 68.3
%&& 28.9
%& 13.9
%\\

100
&Random
% & Similar
&& 63.5%\tiny{2.8}
&  44.2%\tiny{2.9}
& 37.4%\tiny{3.9}
&  89.8%\tiny{1.4}
&  51.5%\tiny{2.9}
&& 65.2%\tiny{1.7}
&& 47.2%\tiny{2.4}
&& 78.6%\tiny{1.6}
& 30.8%\tiny{2.4}
& 15.3%\tiny{0.9}
\\

%\midrule[.05em]

%\modelrandom & 32 & 24.5 & 23.6 && 26.5 &24.6 \\ 
%Diversity
%& Random
%&& 
%& 
%& 
%&  
%&  
%&& 
%&& 
%& 
%&& 
%& 
%
%\\ 

% 100
% &Diversity
% & Similar
% && 67.6%\tiny{2.3}
% & 45.5%\tiny{1.2}
% & 42.9%\tiny{2.2}
% & 92.5%\tiny{1.0}
% & 59.8%\tiny{1.6} 
% && 68.2%\tiny{1.5}
% && 49.2%\tiny{1.2}
% & 82.6%\tiny{1.6}
% && 32.4%\tiny{2.2}
% & 16.3%\tiny{0.7}
% \\

%\midrule[.05em]
%
%\Votek
%& Random
%&& 61.4
%& 38.6
%& 34.8
%&  82.2
%&  51.5
%&& 62.5
%&& 35.6
%& 71.2
%&& 30.4
%& 15.3
%
%\\ 

100
&\Votek
% & Similar
&& \textbf{70.7}%\tiny{1.6}
& \textbf{53.0}%\tiny{1.8}
& \textbf{47.3}%\tiny{2.8}
&  \textbf{93.4}%\tiny{0.9}
&  \textbf{55.5}%\tiny{1.6}
&& \textbf{70.7}%\tiny{1.0}
&& \textbf{51.4}%\tiny{1.8}
&& \textbf{82.8}%\tiny{0.9}
& \textbf{33.6}%\tiny{1.8}
& \textbf{17.2}%\tiny{0.6}
\\

100
% &\multicolumn{2}{c}{$\Delta$ Improvement}
& $\Delta$ Absolute gain
&& \textblue{+7.2}
& \textblue{+8.8}
& \textblue{+9.9}
&  \textblue{+3.6}
&  \textblue{+4.0}
&& \textblue{+5.5}
&& \textblue{+4.2}
&& \textblue{+4.2}
& \textblue{+2.8}
& \textblue{+1.9}
\\

\midrule[.05em]

18
&Random
% & --
&& 59.6%\tiny{6.2}
& 39.8%\tiny{5.4}
& 36.7%\tiny{5.1}
& 77.6%\tiny{5.2}
& 50.4%\tiny{4.2}
&& 62.5%\tiny{4.6}
&& 33.6%\tiny{7.6}
&& 62.4%\tiny{4.0}
& 29.8%\tiny{2.8}
& 13.6%\tiny{1.1}
\\

%\midrule[.05em]

%\modelrandom & 32 & 24.5 & 23.6 && 26.5 &24.6 \\ 
%Diversity
%& Random
%&& 
%& 
%& 
%&  
%&  
%&& 
%&& 
%& 
%&& 
%& 
%
%\\ 

% 18
% &Diversity
% & --
% && 61.4%\tiny{4.8}
% & 39.2%\tiny{4.6}
% & 38.3%\tiny{4.0}
% & 82.9%\tiny{4.3}
% & 57.2%\tiny{3.1} 
% && 64.8%\tiny{3.7}
% && 35.6%\tiny{5.2}
% & 67.3%\tiny{2.8}
% && 31.6%\tiny{2.5}
% & 14.2%\tiny{0.9}
% \\

%\midrule[.05em]
%
%\Votek
%& Random
%&& 61.4
%& 38.6
%& 34.8
%&  82.2
%&  51.5
%&& 62.5
%&& 35.6
%& 71.2
%&& 30.4
%& 15.3
%
%\\ 

18
&\Votek
% & --
&& \textbf{64.2}%\tiny{4.6}
& \textbf{47.6}%\tiny{2.8}
& \textbf{41.0}%\tiny{3.7}
& \textbf{87.1}%\tiny{3.0}
& \textbf{54.3}%\tiny{2.4}
&& \textbf{67.4}%\tiny{3.3}
&& \textbf{42.8}%\tiny{4.2}
&& \textbf{72.5}%\tiny{2.6}
& \textbf{32.3}%\tiny{1.9}
& \textbf{15.2}%\tiny{0.8}
\\

18
% &\multicolumn{2}{c}{$\Delta$ Improvement}
& $\Delta$ Absolute gain
&& \textblue{+4.8}
& \textblue{+7.8}
& \textblue{+4.3}
&  \textblue{+9.5}
&  \textblue{+3.9}
&& \textblue{+4.9}
&& \textblue{+8.8}
&& \textblue{+9.9}
& \textblue{+2.5}
& \textblue{+1.6}
\\

\bottomrule[.1em]
\end{tabular}
\caption{In-context learning results with randomly-selected and \votek \firststep methods on all 10 datasets, with an annotation budget of 100 or 18. 
There is no prompt retrieval step when only 18 samples are annotated since all annotated samples can fit into the in-context learning model's input. 
% We report the standard deviations using subscripts. 
Across the board, \firststep with \votek substantially outperforms the randomly-selected annotation baseline for in-context learning.
% \tao{Hongjin and Chen: TODO add \votek variances. ICL is sensitive to in-context example selection, and \votek reduces ICL variances and makes it more robust and stable.}.
Further, \votek largely reduces the variance over three trials (see the min and max results in Appendix~\ref{app:main-results}), making in-context learning more stable. 
% \chen{Add \votek variances for the 18-example experiment.}
% For random \firststep, we ran every experiment three times and report the average as well as the standard deviation.
}
%We report the standard deviations in Appendix \ref{}, but they are generally small, compared to the improvements from \votek seleection and similarity-based retrieval.
\label{tab:main_results}
\end{table}

%\chen{todo: absolute gain}

%\paragraph{Random Retrieval}
%Similar to random sample selection, we also experiment with a random baseline for the prompt retrieval step as well.
%This quantifies the effect of similarity-based prompt retrieval.

Seen in Table \ref{tab:main_results} are our results from all 10 diverse datasets with the annotation budgets of $|\mathcal{L}| \in \{18, 100\}$.
18 is chosen so that all annotated examples can be fit to the prompt for the language models without prompt retrieval.
Over all datasets, \votek \firststep outperforms the random baseline by a large margin (5.2\% absolute gain and 11.4\% relative gain on average) when the annotation budget is 100.
Even when only 18 examples are annotated and fixed as the in-context examples for all testing instances (no prompt retrieval step), in-context learning with \votek still improves the randomly-selected annotation baseline (5.8\% absolute gain and 12.9\% relative gain on average).
% \footnote{We notice that the performance improvement on NQ and XSum tasks is relatively smaller because the performance impact of in-context examples is much smaller on the two tasks as shown in other recent work \tao{cite}.}.
Particularly noteworthy is that in-context learning with 18 examples selected by \votek achieves higher performance than the one with 100 randomly selected examples on 6 out of 10 tasks.
%, with 5-10x less annotation cost. 
Moreover, \votek is a deterministic \firststep method, conditioned on a set of unlabeled samples. Therefore, the variance of \votek  comes solely from how the unlabeled samples are collected, largely improving the robustness of in-context learning. 
% \tao{add ICL more robust, reducing variances}
We therefore recommend that researchers and practitioners use \firststep (e.g., our \votek method) to better benefit from the few-shot learning capability of large language models with stability.
Our later experiments will also illustrate that \votek consistently outperforms alternative \firststep methods (\S\ref{subsec:sample-selection-methods}).

\section{Analysis}\label{sec:analysis}
Our extensive experiments demonstrated that \firststep is important for the success of in-context learning. 
%\chen{Since random retrieval is removed from the main table, we may only claim the importance of sample selection so far? }
Here we conduct detailed analysis to provide further guidance for researchers and practitioners of few-shot in-context learning.
%Specifically we first compare our in-context learning method with state-of-the-art finetuning methods (\S\ref{sec:in-context_finetuning}).
%We analyze in-context learning from a variety of perspectives: varying language model sizes (\S\ref{sec:lm_sizes}), sample efficiency (\S\ref{sec:sample_efficiency}), the test data domain (\S\ref{sec:domain_shift}), and the diversity and representativeness of selected annotated samples (\S\ref{sec:div_rep_analysis}).
We analyze \firststep for in-context learning from a variety of perspectives: comparisons to finetuning methods (\S\ref{sec:in-context_finetuning}), varying language model sizes (\S\ref{sec:lm_sizes}), test data domain shifts (\S\ref{sec:domain_shift}), prompt retrieval methods (\S\ref{sec:random_retrieval}), and alternative \firststep methods (\S\ref{subsec:sample-selection-methods}). % , and the diversity and representativeness of selected annotated samples (\S\ref{sec:div_rep_analysis}).
%We finally explore an iterative selection approach that is commonly used in active learning \citep{settles.tr09}.

%\subsection{Sample Efficiency of In-Context Learning}
%\label{sec:sample_efficiency}
%In this section, we explore the required number of instances for different settings to achieve similar performance.
%In general, in-context learning with vote-150-prob selection in the first phase is highly sample-efficient.
%For example, in HellaSwag, to achieve similar performance as in-context learning with vote-150-prob selection, the best finetuning setting will require 10 times annotated examples, while in-context learning with randomly-selected annotation will require 8 times.
%This implies in-context learning with sample selection in the first phase helps to significantly save annotation cost, without sacrificing model performance.
%

%\subsection{Better Paradigm}

\subsection{In-Context Learning vs.\ Finetuning}
\label{sec:in-context_finetuning}

\begin{figure}[h!]
\centering
    \includegraphics[width=\textwidth]{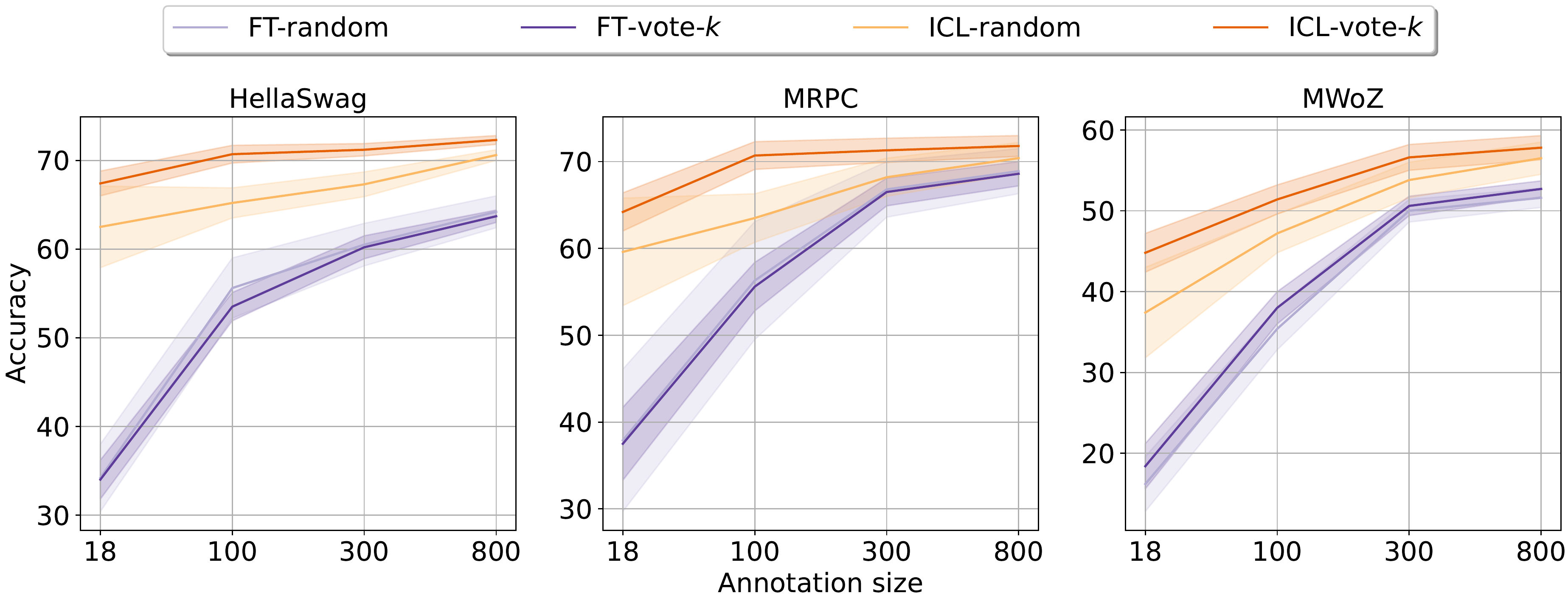}
\caption{Comparisons between the in-context learning and finetuning paradigms over varying annotation budgets on three representative tasks: HellaSwag commonsene reasoning, MRPC paraphrase detection, and MWoZ dialogue state tracking.
%(more experiments over 10 datasets in Tab.~\ref{tab:main_results}).
%\jungo{the text in the brackets is unclear. Cut?}
%(i.e., the size of the labeled samples, $|\mathcal{L}|$)
% Four configurations are presented: finetuning with examples that are randomly selected (FT-random) or selected by our \firststep \votek method (\S\ref{sec:sample_selection_method}; FT-select) and in-context learning with randomly-selected annotation (ICL-random) and \votek selection (ICL-select).
Four configurations are presented: finetuning with examples that are randomly selected to annotate (FT-random) or selected by our \votek \firststep method (\S\ref{sec:sample_selection_method}; FT-vote-\textit{k}) and in-context learning with randomly-selected annotation (ICL-random) or \votek selection (ICL-vote-\textit{k}).
% Two configurations are presented: in-context learning (GPT-J) with examples that are \textit{randomly} selected or selected by our \textit{\votek} \firststep method (\S\ref{sec:sample_selection_method}).
%\jungo{as commented in Fig 1, maybe we should stick to \votek and label it that way?}
%Experimental details are discussed in 
See \S\ref{sec:in-context_finetuning} for experimental details.
% Unlike finetuning paradigm, \firststep largely improves the in-context learning performance compared to randomly-selected annotation even when the annotation budget is 18.
Selective annotation largely improves the in-context learning performance compared to randomly-selected annotation even when the annotation budget is 18.
In-context learning with wisely-selected labeled samples is a much better few-shot practice than a strong finetuning method.
% \tao{update OPT-178b results}
% \chen{figure: vote-k -> \votek}
% \tao{to add: 0. more general overall caption 0.5 hellawag/mrpc gpt3; gpt3 1. fully ft performance; 2. icl retrieving from full training; 3. adding a conclusion in the caption. 4. mentioning which model is used for ICL and fine-tuning. 5. updating the plot for final ICL models. 6. one single horizontal legend; axis labels. 7. mention we did 10 datasets. 8. task -> category}
% \rui{0. I still believe it is very important to show figure 1 right here, but let's provide enough context for readers to understand it. 1. Let's flip the order of methods in legend to align with the curves. 2. One copy of legend is fine outside of the three plots. 3. Let's stretch the horizontal axis to fit textwidth? 4. Is it better to use the same vertical axis range? 5. Is it possible to use orange color for our methods as orange is more noticeable? 6. The current figure/caption doesn't show which language models we use. Is it GPT-3 or something else?}
}
\label{fig:icl-vs-ft}
\end{figure}

As discussed earlier, in-context learning is an alternative learning paradigm to conventional finetuning.
Through the lens of our two-step framework, we observed that \firststep and prompt retrieval are key to the success of in-context learning.
A new question now arises: how does in-context learning compare with finetuning under limited annotation budgets? 
We empirically compare the two paradigms in this section. 
% \tao{Hongjin and Chen TODO: compare to fully finetuning results, and based on icl results retrieving from full training, 10k is not necessary, the marginal gain is very small by annotating large data for ICL.}.

We experiment with three representative tasks: MRPC (classification), HellaSwag (multiple-choice), and MWoZ (dialogue).
Strong, state-of-the-art pretrained models are used for finetuning: large-sized RoBERTa \citep{Liu2019RoBERTaAR} for MRPC and HellaSwag and DS2-T5 \citep{Shin2022DialogueSA} for MWoZ.
In-context learning uses GPT-J for MRPC, GPT-J and OPT 175B (Fig~\ref{fig:hellaswag_opt}) for HellaSwag, and Codex-davinci-002 for MWoZ.
Note that we do not aim to conduct head-to-head comparisons with exactly the same pretrained model; finetuning a large left-to-right language model (e.g., GPT-J and GPT-3) is computationally (and thus financially) infeasible in many cases.
Here we examine the two paradigms from the practical perspective and benefit from the advantage of in-context learning, which requires no parameter updates of massive language models.

Fig.\ \ref{fig:icl-vs-ft} compares the two paradigms across varying annotation sizes ($\{18, 100, 300, 800\}$).
%by a large margin.
% \jungo{Should we move Fig.\ 1? Too far? But I understand showing Fig.\ 1 early is great. Any better alternatives?)}
% The smallest budget of size 18 is chosen so that all of them can be fit as input to the language model as in-context examples.
% Similarity-based prompt retrieval is applied to in-context learning in the other two cases.
Over all three tasks, we observe that in-context learning with \votek selection outperforms the finetuning performance of state-of-the-art pretrained language models. 
Specifically, we find that to achieve similar performance to \votek with $|\mathcal{L}| = $ 18 or 100, finetuning requires 1000 annotated examples for HellaSwag and 800 for MWoZ (\textbf{10-100$\times$ annotation cost}).
Note that the in-context learning performance usually converges when 100 or 300 examples are carefully selected and annotated, suggesting that a large annotated dataset is unnecessary for in-context learning to achieve strong performance.
Interestingly, \firststep helps in-context learning, but \emph{not} finetuning.
This result is consistent with recent work showing that many active learning algorithms perform similarly to random baseline, when pretrained language models are finetuned \citep{karamcheti-etal-2021-mind,darcy2022limitations}.
They proposed that it might be due to outliers and the instability of finetuning on a limited number of annotated samples. 
% \tao{Chen and Hongjin TODO: consider adding our own results}
We hypothesize that in-context learning \textit{with similarity-based prompt retrieval} is more robust to outliers and small annotation sizes because only the most similar examples are retrieved for each test instance. 
We find two pieces of evidence for this hypothesis. First,  \S~\ref{sec:random_retrieval} shows that \textit{random} (as opposed to similarity-based) prompt retrieval does not benefit from \votek \firststep. Second, in Appendix~\ref{app:remove_outlier_FT}, we show that explicitly removing outliers also helps finetuning to benefit from \votek. %\jungo{what is this early experiment referring to? preliminary experiments?} \chen{todo: finalize with Hongjin.}
% Consistent with these prior works, we also explored \firststep methods other than our \votek algorithm (e.g., BALD, \citealp{bald,baldmc}; BatchBALD, \citealp{kirsch2019batchbald}), but we saw no substantial improvement.
% Moreover, \votek with an annotation budget of 800 (and with the subsampling of 3k samples in Section~\ref{sec:datasets_tasks}) achieves similar performance to in-context learning with the entire training data being annotated without the subsampling. This result reveals that in-context learning is very annotation-efficient under our two-step framework. \chen{The last observation is based on the file ``images/with\_full.pdf''. Not sure if to add it. } \chen{Removed. I misunderstood that figure. }

\subsection{Language Models with Various Sizes}
\label{sec:lm_sizes}
\begin{figure}[h!]
\centering
    \includegraphics[width=0.98\textwidth]{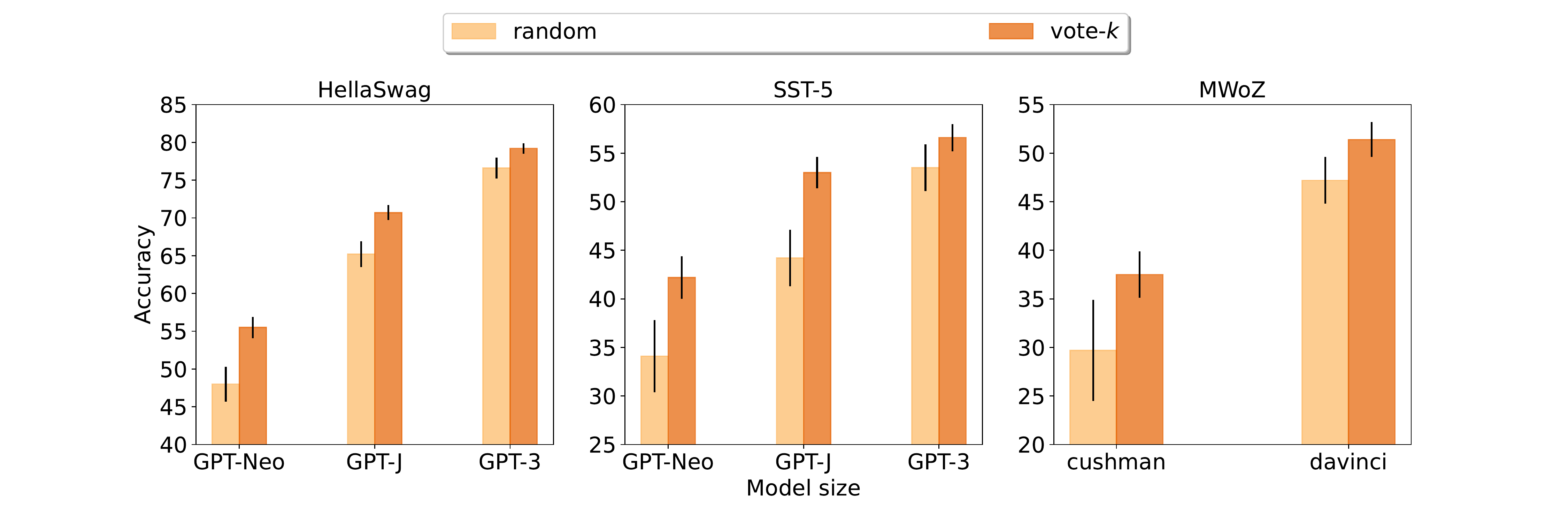}
\caption{
Comparisons of various models with 100 annotated examples. 
\Votek \firststep consistently improves in-context learning with pretrained language models of varying sizes. 
%\chen{Maybe rerank: from smaller models to larger models. }
% \tao{maybe 18}
% \chen{figure: vote-k -> \votek}
}
\label{fig:model_size}
\end{figure}

% So far, we mainly experimented with GPT-J with 6B parameters due to our computational budget. 
Fig.\ \ref{fig:model_size} shows performance with varying sizes of language models (GPT-Neo 2B, \citealp{gpt-neo}; GPT-J 6B, \citealp{gpt-j}; GPT-3, \citealp{gpt3}) on %\nascomment{this dataset is not in figure 3!  mismatch} 
HellaSwag commonsense reasoning, SST-5 sentiment analysis, and MWoZ dialogue state tracking. 
% In all cases, we use similarity-based prompt retrieval (\S\ref{sec:prompt_retrieval}).
In general, when a smaller model is used, the performance gap between random and \votek selection is larger.
In the HellaSwag task, \votek outperforms randomly-selected annotation by 7.5\% using GPT-Neo, but only 2.6\% using GPT-3.
Nonetheless, we see consistent performance gains from \votek selection over varying sizes.
% \tao{show GPT-J/XNeo (5x smaller model size) with wisely selected examples outperforms GPT-3 with random annotated examples?}

% \begin{table*}[h]\small
% \centering
% % \begin{tabular}{P{1.7cm}P{1.2cm}P{1.2cm}P{1.0cm}P{1.2cm}P{1.2cm}P{1.2cm}P{1.1cm}P{1.1cm}P{1.1cm}}
% \begin{tabular}{@{}ccccccc@{}}
% \toprule
%  % Model
%  & \multicolumn{2}{c}{GPT-3 (175B)} & \multicolumn{2}{c}{GPT-J (6B)} & \multicolumn{2}{c}{GPT-Neo (2B)}   \\
%  \cmidrule(lr){2-3} \cmidrule(lr){4-5} \cmidrule(l){6-7}
%  Selection & Random & \Votek & Random & \Votek & Random & \Votek   \\
% \midrule
% HellaSwag & 76.6\tiny{1.4} & 79.2 & 65.2\tiny{1.7} & 70.7 & 53.0\tiny{2.3} &  55.5  \\
% SST-5 & 53.5\tiny{2.4} & 56.6 & 44.2\tiny{2.9} & 53.0 & 34.1\tiny{3.7} &  42.2  \\
% \bottomrule
% \end{tabular}
% \begin{tabular}{@{}ccc@{}}
% \toprule
%  % Model
%  & \multicolumn{2}{c}{MWoZ} \\
%  \cmidrule(lr){2-3}
%  Model & Random & \Votek \\
% \midrule
% Davinci & 47.2\tiny{2.4} & 51.4  \\
% cushman & 29.7\tiny{5.2} & 37.5 \\
% \bottomrule
% \end{tabular}
% \caption{
% \label{tab:model_size}
% Comparisons of various models with 100 annotated examples.
% \tao{consider plotting and add a takeaway \rui{Plot is better as we can visual model size in horizontal axis.}} \Votek \firststep consistently improves in-context learning with pre-trained language models of varying sizes. \chen{Maybe rerank: from smaller models to larger models. }
% }
% \end{table*}

\subsection{Effects of Domain Shift}
\label{sec:domain_shift}
Recent work observed that when a large, pretrained language model is finetuned, the performance gain from active learning is limited \citep{darcy2022limitations}, but it can be larger if there is a domain shift between training and evaluation \citep{tamkin22}.
We have demonstrated that \firststep consistently improves in-context learning, but here we explore cases of domain shifts.

\begin{table}[h!] %[12]{r}{9.5cm}
\centering
% \vspace{-5mm}
\begin{tabular}{@{} cc   m{0.001em}  cc  m{0.001em} cc @{}}
\toprule[.1em]

 \multicolumn{2}{c}{\textbf{Method}}
&& \multicolumn{2}{c}{\textbf{CivilComments}}
&& \multicolumn{2}{c}{\textbf{Amazon}}
\\
\cmidrule(r){1-2}
\cmidrule(lr){4-5}
\cmidrule(l){7-8}

$|\mathcal{L}|$ & Selection
&& Random & Domain
&& Random & Domain
\\

\midrule[.1em]

%Random
%& Random
%&&  72.7 
%& 65.4
%&& 48.6
%& 24.7
%\\

100 & Random
&& 73.8%\tiny{4.2}
& 66.8%\tiny{6.4}
&& 50.3%\tiny{4.8}
& 30.7%\tiny{6.2}
\\

%\Votek
%& Random
%&& 75.8
%& 72.3
%&& 51.2
%& 32.8 
%
%\\ 
100 &
\Votek
&& \textbf{79.3}%\tiny{2.6}
& \textbf{76.7}%\tiny{3.2}
&& \textbf{56.3}%\tiny{3.1}
& \textbf{39.0}%\tiny{3.6}
\\

100 & $\Delta$ Absolute gain
&& \textblue{+5.5}
& \textblue{+9.9}
&&  \textblue{+6.0}
& \textblue{+8.3}
\\

\bottomrule[.1em]
\end{tabular}
\caption{Effects of domain shift. Random splits and domain splits are compared \citep{wilds}.
%In both cases, we annotate 100 examples only ($|\mathcal{L}|=100$).
% \tao{add \votek variance}
% \rui{Maybe use two tables?} \tao{agree! even just use plots}
}
%We follow \citet{margatina2021active} and quantify the properties of selected samples using different methods.
\label{tab:domain_shift}
\end{table}

Following \citet{tamkin22}, we use two natural language datasets from the WILDS benchmark \citep{wilds}: \textbf{CivilComments} (toxicity classification; \citealp{civilcomments}) and \textbf{Amazon} (review classification; \citealp{ni-etal-2019-justifying}).
Each comes with both a random split and a domain split: the former splits data randomly and the latter is based on the domains (demographic identities for CivilComments and users for Amazon), simulating cases where a model is deployed in a new scenario unseen during annotations.
Similar to \S\ref{sec:results}, we conduct experiments with GPT-J under two settings: random/\votek \firststep, followed by similarity-based prompt retrieval. Both \firststep and prompt retrieval are conducted on the source domain. 

Tab.~\ref{tab:domain_shift} shows our results. 
We see that the gain from \votek is more pronounced under the domain splits: e.g., \textblue{9.9} vs.\ \textblue{5.5} accuracy point improvements on CivilComments. 
This suggests that \firststep and prompt retrieval are particularly crucial when there is a domain shift in the evaluation data, as in many realistic scenarios \citep{wilds,longpre2022active}.

\subsection{Random Prompt Retrieval}
\label{sec:random_retrieval}

\begin{table}[!th] %[12]{r}{8.2cm}
\centering
% \addtolength{\tabcolsep}{-4.0pt}
% \vspace{-15mm}
\begin{tabular}{@{} ccc   m{0.001em}  ccc }
\toprule[.1em]

 \multicolumn{3}{c}{\textbf{Method}}
&& 
 \multicolumn{3}{c}{\textbf{Dataset}}
\\
\cmidrule(lr){1-3}
\cmidrule(lr){5-7}

$|\mathcal{L}|$ & Selection
& Retrieval
&&  HellaSwag
& SST-5 
& MWoZ
\\

\midrule[.1em]

%Random
%& Random
%&&  72.7 
%& 65.4
%&& 48.6
%& 24.7
%\\

100 & \Votek
& Similar
&& \textbf{70.7}%\tiny{1.0}
& \textbf{53.0}%\tiny{1.8}
& \textbf{51.4} %\tiny{1.8}
\\

100 & Random
& Similar 
&& 65.2
& 44.2
& 47.2
\\

100 & \Votek
& Random
&&  62.5%\tiny{2.1}
& 41.6%\tiny{2.2}
& 35.6%\tiny{4.6}
\\

100 & Random
& Random
&& 63.2%\tiny{2.5} 
& 40.6%\tiny{3.2}
& 43.8%\tiny{5.2}
\\

\bottomrule[.1em]
\end{tabular}
\caption{Comparison of random and similar prompt retrieval. 
Random retrieval fails to benefit from diverse and representative annotated examples from \votek. 
%\tao{Hongjin check: 18? also >100?}
%\chen{Copied random-similar here, as Sewon suggested. }
%$|\mathcal{L}|=100$.
% \tao{\votek variance}
}
\label{tab:random_retrieval}
\end{table}

We have performed similarity-based prompt retrieval so far.
Here we experiment with a random baseline for the prompt retrieval step to quantify the effect of prompt retrieval (Tab.~\ref{tab:random_retrieval}).
Interestingly, when random prompt retrieval is performed, \votek does not necessarily improve upon the randomly-selected annotation baseline: e.g., 62.5 vs.\ 63.2 on HellaSwag and 35.7 vs.\ 43.8 on MWoZ.
This suggests that random prompt retrieval fails to benefit from diverse, representative 100 samples, selected by \votek \firststep. 
%When randomly-selected annotation is applied, on the other hand, similarity-based retrieval still outperforms random retrieval on all datasets: e.g., 44.2 vs.\ 37.6 on SST-5. Nonetheless, it substantially underperforms the combination of \votek selection and similarity-based retrieval.
Combining \firststep and prompt retrieval is thus crucial for the success of in-context learning.

%For least confidence, even with iterative selection, it fails to outperform vote-150. 
%The possible reason is the lack of diversity of selected examples in very iteration.

\subsection{Alternative \FirstStep Methods}

\begin{table*}[h!]
\centering
% \begin{tabular}{P{1.7cm}P{1.2cm}P{1.2cm}P{1.0cm}P{1.2cm}P{1.2cm}P{1.2cm}P{1.1cm}P{1.1cm}P{1.1cm}}
\begin{tabular}{@{}ccccccc@{}}
\toprule
 & Random & MFL & Diversity & Least-confidence & Fast \votek &  \Votek   \\
\midrule
HellaSwag & 65.2%\tiny{1.7} 
& 66.5%\tiny{1.7} 
& 68.2%\tiny{1.5} 
& 68.4%\tiny{1.0} 
& 69.5%\tiny{1.2} 
&  \textbf{70.7}%\tiny{1.0}  
\\
SST-5 & 44.2%\tiny{2.9} 
& 45.6%\tiny{2.4} 
& 48.5%\tiny{2.2} 
& 46.2%\tiny{1.9} 
& 51.9%\tiny{2.1} 
& \textbf{53.0}%\tiny{1.8}  
\\
MWoZ & 47.2%\tiny{2.4} 
& 48.3%\tiny{2.0} 
& 49.2%\tiny{2.2} 
& 49.4%\tiny{1.8} 
& 50.2%\tiny{2.0}  
& \textbf{51.4}%\tiny{1.8}  
\\
\bottomrule
\end{tabular}
\caption{
\label{all_sample_selection_method}
Comparisons of various selective annotation methods with 100 annotated examples. Performance is averaged over three random trials.
% \tao{add \votek variance.} 
\Votek outperforms all the other \firststep methods.  Fast \votek, a faster version of voke-k without the need for confidence score computations, can achieve similar performance to \votek while being more computationally efficient.
% \tao{to fig?}
}
\end{table*}

% \begin{figure}[h!]
% \centering
%     \includegraphics[width=0.98\textwidth]{images/selection_methods.pdf}
% \caption{Comparisons of various selection methods with 100 annotated examples. Performance is averaged over three random trials.
% % \tao{add \votek variance.} 
% \Votek outperforms all the other \firststep methods.  Fast \votek, a faster version of voke-k without the need for confidence score computations, can achieve similar performance to \votek while being more computationally efficient. 
% }
% \label{fig:hellaswag_opt}
% \end{figure}

\label{subsec:sample-selection-methods}
Here we explore four additional methods for \firststep: 
\begin{compactitem}
\item \textbf{Maximizing facility location} (MFL; \citealp{Lin2009HowTS}) aims at optimizing the representativeness of the selected samples. Since this objective satisfies the submodular objective, maximization can be approximated via a greedy algorithm (see Appendix \ref{sec:Submodularity-Based Sample Selection}).
\item \textbf{Diversity} focuses on maximizing the diversity of the embeddings for selected examples in the first step (Appendix \ref{sec:Embedding Diversity}).
\item \textbf{Least-confidence} \citep{lewis1994sequential} iteratively adds least-confident examples to the annotated set.
\item \textbf{Fast \votek} is a fast, efficient alternative to our \votek method (\S\ref{sec:sample_selection_method}) that does not use confidence scores.
It picks $M$ samples with the largest \votek scores.
It avoids using the pretrained language model to compute a confidence score for every instance, resulting in a 10+ times speedup.
% \jungo{Hongjin, can you mention how fast fast \votek is compared to \votek??}
\end{compactitem}

%\tao{Chen TODO: mention other methods that don't have hyperparameters and add Submodularity vs. \Votek discussion here. the efficiency of each method}
Notice that MFL, diversity, and least-confidence do not have hyperparameters other than the annotation budget. 
As shown in Tab.\ \ref{all_sample_selection_method}, \votek outperforms all the other methods.
%Diversity is efficient and straightforward.
%It outperforms randomly-selected annotation by large margin, and is close to vote-150.
It is noteworthy, however, that fast \votek can achieve similar performance to \votek. 
Fast \votek is thus an attractive method for researchers and practitioners with a limited computational budget.
%\tao{too add more comparison details and findings}
Like \votek, MFL also optimizes representativeness and Diversity also optimizes diversity. In particular, MFL defines representativeness as a sum over distances from the selected examples to all other examples, and Diversity defines diversity as the distances between selected examples. Since they do not significantly outperform randomly-selected annotation, we conjecture that jointly optimize diversity and representativeness is needed for \firststep. 
Moreover, the way \votek defines and diversity are also different from the baselines: \votek defines representativeness as the number of neighbors during similarity-based prompt retrieval, which is effectively tailored to in-context learning; \votek directly optimizes for the diversity of selected samples using the in-context learning model's prediction confidence. 
%\tao{more analysis experiments?} 

\section{Related Work}\label{sec:related}
\paragraph{In-Context Learning} 
In-context learning with large language models has recently received an increasing amount of interest, partly due to its flexibility and sample efficiency \citep{prompt_survey}.
Several recent works proposed methods to improve in-context learning in many aspects: e.g., meta-training \citep{chen-etal-2022-meta,min-etal-2022-metaicl}, task instructions \citep{efrat20,mishra-etal-2022-cross,Wei2021FinetunedLM,sahn22}, or task formulation \citep{holtzman-etal-2021-surface,calibrate,min-etal-2022-noisy}.
In this paradigm, the choice of in-context (i.e., demonstration) examples has been shown crucial \citep{liu-etal-2022-makes,rubin2022,lu-etal-2022-fantastically}, while recent work raised questions as to the degree to which correct labels are necessary \citep{min2022rethinking}.
This work proposes an annotation-efficient in-context learning framework by focusing on the choice of examples and its implications on the annotation cost.

% Recent work \citep{liu-etal-2022-makes,rubin2022} improved the performance by retrieving similar examples to each test instance in some embedding space (e.g., BM25, \citealp{bm25}; RoBERTa, \citealp{Liu2019RoBERTaAR}; Sentence-BERT, \citealp{reimers-gurevych-2019-sentence}).

\paragraph{Active Learning}
%\tao{Chen TODO: finalize and edit this section}
Active learning aims to enable machine learning models to achieve similar or greater performance with fewer labeled training instances~\citep{CohnAL94,settles.tr09}.
Our \firststep step for in-context learning shares the same goal of reducing the annotation cost.
Most active learning methods involve iterative parameter updates (e.g., \citealp{active_image_recog,kasai-etal-2019-low}), which are computationally expensive for large language models used in in-context learning.
%As natural language processing tasks typically require a large number of annotated training examples, much prior work explored active learning algorithms to reduce the annotation cost \interalia{thompson99,hwa-2000-sample,tang-etal-2002-active}.
Similar to our \votek algorithm, \citet{Lin2009HowTS} used the facility location 
% \nascomment{
%here and in the earlier discussion of Lin and Bilmes, I'm worried that you're packaging things together in a misleading way.  IIRC, Lin and Bilmes use a particular objective to frame the selection problem, and their algorithm exploits the fact that that particular objective has a submodular property.  I don't remember enough to reconstruct what the objective was, but I am worried that you are conflating the submodular property they exploited with their general approach.  what did they call their method?  
% second, I wonder if our objective (or the simplified version in the ``fast'' \votek variant, which is easier to think about, is submodular and hence amenable to being used with their algorithm. } 
objective to optimize representativeness.
%Their algorithm tends to select outliers \citep{karamcheti-etal-2021-mind}, resulting in worse performance than \votek (\S\ref{subsec:sample-selection-methods}).
% \jungo{Check, Hongjin}
% \chen{This point seems to contradict to our claim that in-context learning is robust to outliers. }
%However, classical active learning algorithms are often prohibitively expensive for large-scale models.
%Consequently, many acceleration methods have been proposed ~\citep{sener2018active, coleman2020selection, yoo2019learning}.
We observed that this objective largely underperforms \votek for in-context learning, probably due to the fact the \votek (1) is effectively tailored to the prompt retrieval step of in-context learning and (2) directly optimizes the diversity of selected samples (see \S\ref{subsec:sample-selection-methods}). 
More recently, the effectiveness of active learning has been questioned when large-scale pretrained models are finetuned for various tasks \citep{karamcheti-etal-2021-mind,darcy2022limitations}.
Our experiments (\S\ref{sec:experiments}) showed that \firststep helps reduce the annotation cost of in-context learning, departing from the recent observations on finetuning with active learning.
We hypothesize that it is because in-context learning with similarity-based prompt retrieval is more robust to outliers since each test instance only retrieves its most similar examples. 
This is supported by \S~\ref{sec:random_retrieval}, where \textit{random} prompt retrieval does not benefit from \firststep.

\section{Conclusion}\label{sec:conclusion}
Much recent work illustrated the ability of large language models to adapt to new tasks simply from a few demonstration examples.
We presented in-depth studies on the implications of this ability for dataset annotation through the lens of \firststep and introduced an annotation-efficient practice. The best \firststep method explored in this paper, our \votek method, selects diverse, representative examples to annotate. 
In terms of the task performance, \votek improves the performance on 10 diverse tasks by a large margin.
%In terms of the annotation cost,
Moreover, \votek \firststep yields similar performance to state-of-the-art supervised finetuning with 10-100$\times$ less annotation cost. 
We further show that the effectiveness of \votek is consistent with different language model sizes and domain shifts between training and test data.
We hope that our findings will help researchers and practitioners efficiently design new natural language tasks and beyond.
\section*{Acknowledgements}
We thank Sewon Min, Pradeep Dasigi, Yanda Chen, Yushi Hu, Alisa Liu, and the ARK group at UW for their helpful feedback on this work.

\bibliography{custom}
\bibliographystyle{iclr2022_conference}

\clearpage
\appendix
\begin{appendices}
\section{Datasets and Tasks}
\label{sec:datasets_all}

\begin{table*}
\small
\begin{tabular}{C{1.5cm}|C{2.5cm}|L{8.5cm}}
\toprule
Dataset & Task & Examples \\ 
\midrule[.005em]
HellaSwag & Commonsense \par Reasoning &
\textcolor{gray}{A woman is outside with a bucket and a dog. The dog is running
around trying to avoid a bath. She…} \par
\phantom{\cmark}A) rinses the bucket off with soap and blow dry the dog’s head.\par
\phantom{\cmark}B) uses a hose to keep it from getting soapy.\par
\cmark C) gets the dog wet, then it runs away again.\par
\phantom{\cmark}D) gets into a bath tub with the dog.
\\
\midrule[.005em]
MRPC & Paraphrase Detection
&
\textcolor{gray}{Sales rose 37 per cent year-on-year to 1.76bn, beating expectations.
Sales for the quarter beat expectations, rising 37 percent year-on-year to 1.76 billion euros.}\par
$\rightarrow$\ \cmark \ Paraphrase
\\
\midrule[.005em]
SST & Sentiment Analysis & \textcolor{gray}{A warm, funny, engaging film.$\rightarrow$}Positive \par
\textcolor{gray}{Suffers from the lack of a compelling narrative.$\rightarrow$}Negative
 \\
\midrule[.005em]
MWoZ 2.4 & Dialogue State Tracking & \textcolor{gray}{I am looking for ALexender b\&b}\par
\textcolor{gray}{Dialogue state:} alexander bed and breakfast 
\\
\midrule[.005em]
GeoQuery & Semantic Parsing & \textcolor{gray}{What is the area of California?}
\begin{lstlisting}[
          language=SQL,
          aboveskip=-0.1 \baselineskip,
          belowskip=-2.2 \baselineskip,
        ]
SELECT state.area FROM state WHERE state.state_name='california'
\end{lstlisting}
\vspace{-3cm}
\\
\midrule[.005em]
DBpedia & Topic Classification & \textcolor{gray}{The keeled box turtle (Cuora mouhotii syn. Pyxidea mouhotii) is a species of turtle in the family Geoemydidae. It is native to Asia where it occurs in China India Laos Burma Vietnam Thailand and Bhutan. Other common names include keel-backed terrapin and jagged-shelled turtle.}\par
\textcolor{gray}{Topic:} animal 
\\
\midrule[.005em]
MNLI & Natural Language Inference & \textcolor{gray}{The F/A-18-E/F program eliminated over 40 percent of the parts used to build predecessor aircraft to make the design more robust for manufacturing and identified critical manufacturing processes, bringing them under control before the start of production.
The new design with robustness also increased the safety of machines.}\par
$\rightarrow$\ \cmark \ neutral
\\
\midrule[.005em]
RTE & Natural Language Inference & \textcolor{gray}{Judie Vivian, chief executive at ProMedica, a medical service company that helps sustain the 2-year-old Vietnam Heart Institute in Ho Chi Minh City (formerly Saigon), said that so far about 1,500 children have received treatment. The previous name of Ho Chi Minh City was Saigon.}\par
$\rightarrow$\ \cmark \ entailment
\\
\midrule[.005em]
Natural Questions & Open-Domain QA & \textcolor{gray}{when was music first played on the radio}\par
$\rightarrow$\ \cmark \ 1917
\\
\midrule[.005em]
XSUM & Summarization & \textcolor{gray}{Bliss said there was a shortage of neonatal nurses and doctors, and safety standards were not being met.
......
Dr Jenny Calvert, of the Wales Neonatal Network, said they are working to further develop medical training in neonatology to help recruit more trainee doctors.}
\textcolor{gray}{Summary:} Neonatal services across Wales are overstretched and under pressure with the safety of vulnerable babies at risk, according to a charity.
\\
\bottomrule
\end{tabular}
\caption{
All of the 10 datasets with examples used in our experiments.
The 10 datasets span various formations, including classification (SST-5,  \citealp{socher-etal-2013-recursive}; MRPC, \citealp{mrpc}), multiple-choice selection (HellaSwag, \citealp{zellers-etal-2019-hellaswag}), and code/text generation (MWoZ 2.4, \citealp{budzianowski-etal-2018-multiwoz};  GeoQuery, \citealp{geoquery96}; NQ, \citealp{kwiatkowski2019natural}).}
\label{tab:datasets_all}
\end{table*}

\clearpage

\section{Prompt Templates}
\label{sec:prompt_templates}
% Seen in Table are prompt templates that we used for each of the 10 datasets.
\subsection{HellaSwag}\par
\textbf{Input:}\par
\begin{lstlisting}[language={}]
The topic is Grooming dog. Two women attempt to wash two dogs. they get 
in the tub with the dogs and do shampoo, soap, and then rinse the dogs.
......
The topic is Bathing dog. A couple is outside with a bucket and a dog. 
The dog is running around trying to avoid a bath. they
\end{lstlisting}

\vspace{-20pt}

\textbf{Output:}

\begin{lstlisting}[language={}]
get the dog wet, then it runs away again.
\end{lstlisting}

\vspace{-20pt}

\subsection{MRPC}
\textbf{Input:}

\begin{lstlisting}[language={}]
Are the following two sentences 'equivalent' or 'not equivalent'?
This was around the time Congress was debating a resolution granting the 
President broad authority to wage war ..
Within four days , the House and Senate overwhelmingly endorsed a 
resolution granting the president authority to go to war ..
answer:not equivalent
......
Are the following two sentences 'equivalent' or 'not equivalent'?
Kerry last month outlined a U.N. resolution authorizing a military force 
under U.S. command and transferring responsibility to the United Nations 
for the political and humanitarian efforts ..
Kerry outlined last month a UN resolution authorizing a military force 
under US command and transferring responsibility for political and 
humanitarian efforts to the UN ..
answer:
\end{lstlisting}

\vspace{-20pt}

\textbf{Output:}

\begin{lstlisting}[language={}]
equivalent
\end{lstlisting}

\vspace{-20pt}

\subsection{SST5}

\textbf{Input:}

\begin{lstlisting}[language={}]
How do you feel about the following sentence?
the movie 's blatant derivativeness is one reason it 's so lackluster .
answer:negative
......
How do you feel about the following sentence?
the movie 's something-borrowed construction feels less the product of 
loving , well integrated homage and more like a mere excuse for the wan , 
thinly sketched story .
answer:
\end{lstlisting}

\vspace{-20pt}

\textbf{Output:}

\begin{lstlisting}[language={}]
negative
\end{lstlisting}

\clearpage

\subsection{MultiWoz}
\textbf{Input:}

\begin{lstlisting}[language={}]
CREATE TABLE hotel(
  name text,
  ......,
  internet text CHECK (internet IN (dontcare, yes, no))
)
/*
4 example rows:
SELECT * FROM hotel LIMIT 4;
name  pricerange  type  parking book_number_of_days book_day  book_people 
area  stars internet
a and b guest house moderate guest house dontcare 3 friday 5 east 4 yes
......
/*
......
-- Using valid SQLite, answer the following multi-turn conversational 
questions for the tables provided above.
Example #1
[context] hotel-area: west, hotel-stars: 3, hotel-internet: yes
[system] the hobsons house is available in that area .
Q: [user] that sounds like it will work . can i book that for 3 nights
starting wednesday ?
SQL: SELECT * FROM hotel WHERE book_day = wednesday AND book_people = 1 
AND book_number_of_days = 3 AND name = hobsons house;
......
Example #22
[context] hotel-parking: yes, hotel-pricerange: moderate, hotel-type: 
guest house, hotel-stars: 4
[system] there are 9 in the area . i recommend the warkworth house .
Q: [user] can you book that 1 for 4 nights starting on wednesday ?
SQL: SELECT * FROM
\end{lstlisting}

\vspace{-20pt}

\textbf{Output:}
\begin{lstlisting}[language={}]
hotel WHERE book_day = wednesday AND book_number_of_days = 4 AND name = 
warkworth house;
\end{lstlisting}

\vspace{-20pt}

\subsection{GeoQuery}
\textbf{Input:}

\begin{lstlisting}[language={}]
CREATE TABLE "border_info" ("state_name" text, "border" text)
/*
state_name    border
   alabama tennessee
   alabama   georgia
   alabama   florida
*/
......
-- Using valid SQLite, answer the following questions for the tables 
provided above.
-- which state has the longest river
SELECT RIVERalias0.TRAVERSE FROM RIVER AS RIVERalias0 WHERE RIVERalias0.
LENGTH = ( SELECT MAX( RIVERalias1.LENGTH ) FROM RIVER AS RIVERalias1 ) ;
......
-- what is the longest river in the state with the highest point
SELECT
\end{lstlisting}

\clearpage

\textbf{Output:}

\begin{lstlisting}[language={}]
RIVERalias0.RIVER_NAME FROM HIGHLOW AS HIGHLOWalias0 , RIVER AS 
RIVERalias0 WHERE HIGHLOWalias0.HIGHEST_ELEVATION = ( SELECT MAX(
HIGHLOWalias1.HIGHEST_ELEVATION ) FROM HIGHLOW AS HIGHLOWalias1 ) AND
RIVERalias0.TRAVERSE = HIGHLOWalias0.STATE_NAME ORDER BY RIVERalias0.
LENGTH DESC LIMIT 1 ;
\end{lstlisting}

\vspace{-20pt}

\subsection{DBpedia}
\textbf{Input:}
\begin{lstlisting}[language={}]
title: Cupressus funebris; content:  Cupressus funebris (Chinese Weeping
Cypress) is a species of cypress native to southwestern and central 
China. It may also occur naturally in Vietnam.
plant
......
title: Keeled box turtle; content:  The keeled box turtle (Cuora mouhotii 
syn. Pyxidea mouhotii) is a species of turtle in the family Geoemydidae. 
It is native to Asia where it occurs in China India Laos Burma Vietnam
Thailand and Bhutan. Other common names include keel-backed terrapin and
jagged-shelled turtle.
\end{lstlisting}

\vspace{-20pt}

\textbf{Output:}
\begin{lstlisting}[language={}]
animal
\end{lstlisting}

\vspace{-20pt}

\subsection{MNLI}
\textbf{Input:}

\begin{lstlisting}[language={}]
Ideally, the design fixes for the failures should be corrected prior to
manufacturing production units.. Based on that information, is the claim 
The fixes should be addressed before they reach the assembly line if this 
was a smart plan. "True", "False", or "Inconclusive"?
answer:Inconclusive
......
The F/A-18-E/F program eliminated over 40 percent of the parts used to 
build predecessor aircraft to make the design more robust for 
manufacturing and identified critical manufacturing processes, bringing 
them under control before the start of production.. Based on that 
information, is the claim The new design with robustness also increased 
the safety of machines.  "True", "False", or "Inconclusive"?
answer:
\end{lstlisting}

\vspace{-20pt}

\textbf{Output:}
\begin{lstlisting}[language={}]
Inconclusive
\end{lstlisting}

\vspace{-20pt}

\subsection{RTE}
\textbf{Input:}
\begin{lstlisting}[language={}]
After giving nearly 5,000 people a second chance at life, doctors are
celebrating the 25th anniversary of Britian's first heart transplant 
which was performed at Cambridgeshire's Papworth Hospital in 1979..\par
question: The first heart transplant in Britian was performed in 1979.. 
True or False?
answer:True
......
Judie Vivian, chief executive at ProMedica, a medical service company 
that helps sustain the 2-year-old Vietnam Heart Institute in Ho Chi Minh 
City (formerly Saigon), said that so far about 1,500 children have 
received treatment..
question: The previous name of Ho Chi Minh City was Saigon.. True or 
False?
answer:
\end{lstlisting}

\vspace{-20pt}

\textbf{Output:}

\begin{lstlisting}[language={}]
True
\end{lstlisting}

\vspace{-20pt}

\subsection{Natural Question}

\textbf{Input:}
\begin{lstlisting}[language={}]
Write an answer: who invented the radio during the industrial revolution
other
Guglielmo Marconi, 1st Marquis of Marconi
......
Write an answer: when was music first played on the radio
\end{lstlisting}

\vspace{-20pt}

\textbf{Output:}
\begin{lstlisting}[language={}]
other
1917
\end{lstlisting}

\vspace{-20pt}

\subsection{XSUM}
\textbf{Input:}

\begin{lstlisting}[language={}]
Write a short summary
Health Minister Mark Drakeford said the money would be used to improve 
areas of concern, including out-of-hours help and access to psychological 
treatment.
......
money won't get the help they need in a timely fashion," she said.
TL;DR: An extra [Unicode token]7.6m a year will be invested to improve 
mental health services for children and young people in Wales.
......
write a short summary:
Bliss said there was a shortage of neonatal nurses and doctors, and 
safety standards were not being met.
......
Dr Jenny Calvert, of the Wales Neonatal Network, said they are working to
further develop medical training in neonatology to help recruit more 
trainee doctors.
TL;DR:
\end{lstlisting}

\vspace{-20pt}

\textbf{Output:}

\begin{lstlisting}[language={}]
Neonatal services across Wales are overstretched and under pressure with 
the safety of vulnerable babies at risk, according to a charity.
\end{lstlisting}

\clearpage

\section{Detailed Main Results}
\label{app:main-results}

This section provides a detailed version of our main results in Table~\ref{tab:main_results}, where the maximum performance and minimum performances among the three trials are reported. Results are shown in Table~\ref{tab:app-main-results-1} and Table~\ref{tab:app-main-results-2}.

\begin{table}[h!]
\addtolength{\tabcolsep}{-4.7pt}
\centering
\begin{tabular}{@{} cc   m{0.001em}  ccccc m{-0.05mm} @{}}
\toprule[.1em]

 \multicolumn{2}{c}{\textbf{Method}}
&& \multicolumn{5}{c}{\textbf{Classification}}
\\
\cmidrule(lr){1-2}
\cmidrule(l){4-8}

$|\mathcal{L}|$
&Selection
% & Retrieval
&& MRPC & SST-5 & MNLI & DBpedia & RTE
\\

\midrule[.1em]

100
&Random
% & Similar
&& 63.5/66.0/60.5%\tiny{2.8}
&  44.2/47.3/41.8%\tiny{2.9}
&  37.4/41.0/33.2%\tiny{3.9}
&  89.8/91.0/88.3%\tiny{1.4}
&  51.5/53.9/48.4%\tiny{2.9}
\\

100
&\Votek
% & Similar
&& \textbf{70.7}/72.3/69.1%\tiny{1.6}
& \textbf{53.0}/54.7/51.2%\tiny{1.8}
& \textbf{47.3}/50.0/44.5%\tiny{2.8}
&  \textbf{93.4}/94.1/92.6%\tiny{0.8}
&  \textbf{55.5}/57.0/53.9%\tiny{1.6}
\\

\midrule[.05em]

18
&Random
% & --
&& 59.6/64.8/52.7%\tiny{6.2}
& 39.8/46.1/37.1%\tiny{5.4}
& 36.7/40.6/30.9%\tiny{5.1}
& 77.6/82.0/71.9%\tiny{5.2}
& 50.4/53.5/45.7%\tiny{4.1}
\\

18
&\Votek
% & --
&& \textbf{64.2}/67.6/59.0%\tiny{4.6}
& \textbf{47.6}/50.0/44.5%\tiny{2.8}
& \textbf{41.0}/44.5/37.1%\tiny{3.7}
& \textbf{87.1}/90.6/85.2%\tiny{3.0}
& \textbf{54.3}/56.2/51.6%\tiny{2.4}
\\

\bottomrule[.1em]
\end{tabular}
\caption{Main result Table~\ref{tab:main_results} with the mean/max/min results reported across three trials}
\label{tab:app-main-results-1}
\end{table}

\begin{table}[h!]
\addtolength{\tabcolsep}{-4.7pt}
\centering
\begin{tabular}{@{} cc   m{0.001em}  c m{-0.05em}  c m{0.001em}   ccc @{}}
\toprule[.1em]

 \multicolumn{2}{c}{\textbf{Method}}
&& \multicolumn{1}{c}{\textbf{Multi-Choice}}
&& \multicolumn{1}{c}{\textbf{Dialogue}}
&& \multicolumn{3}{c}{\textbf{Generation}}
\\
\cmidrule(lr){1-2}
\cmidrule(lr){4-5}
\cmidrule(lr){6-7}
\cmidrule(l){8-10}

$|\mathcal{L}|$
&Selection
% & Retrieval
&& HSwag
&& MWoZ 
&& GeoQ & NQ & XSum
\\

\midrule[.1em]

100
&Random
% & Similar
&& 65.2/66.4/63.3%\tiny{1.7}
&& 47.2/49.2/44.5%\tiny{2.4}
&& 78.6/80.5/77.3%\tiny{1.7}
& 30.8/32.8/28.1%\tiny{2.4}
& 15.3/16.4/14.8%\tiny{0.9}
\\

100
&\Votek
% & Similar
&& \textbf{70.7}/71.5/69.5%\tiny{1.1}
&& \textbf{51.4}/53.1/49.6%\tiny{1.8}
&& \textbf{82.8}/83.6/82.0%\tiny{0.9}
& \textbf{33.6}/35.2/31.6%\tiny{1.8}
& \textbf{17.2}/17.6/16.4%\tiny{0.7}
\\

\midrule[.05em]

18
&Random
% & --
&& 62.5/66.4/57.4%\tiny{4.6}
&& 33.6/39.5/25.0%\tiny{7.6}
&& 62.4/65.2/57.8%\tiny{4.0}
& 29.8/31.6/26.6%\tiny{2.8}
& 13.6/14.5/12.5%\tiny{1.0}
\\

18
&\Votek
% & --
&& \textbf{67.4}/71.1/64.8%\tiny{3.3}
&& \textbf{42.8}/47.7/40.2%\tiny{4.2}
&& \textbf{72.5}/74.2/69.5%\tiny{2.6}
& \textbf{32.3}/33.6/30.1%\tiny{1.9}
& \textbf{15.2}/16.0/14.5%\tiny{0.8}
\\

\bottomrule[.1em]
\end{tabular}
\caption{Main result Table~\ref{tab:main_results} with the mean/max/min results reported across three trials.}
\label{tab:app-main-results-2}
\end{table}

\section{Evaluate HellaSwag on OPT-175B model}
Here we show that \votek also improves model performance for OPT-175B
\begin{figure}[h!]
\centering
    \includegraphics[width=0.45\textwidth]{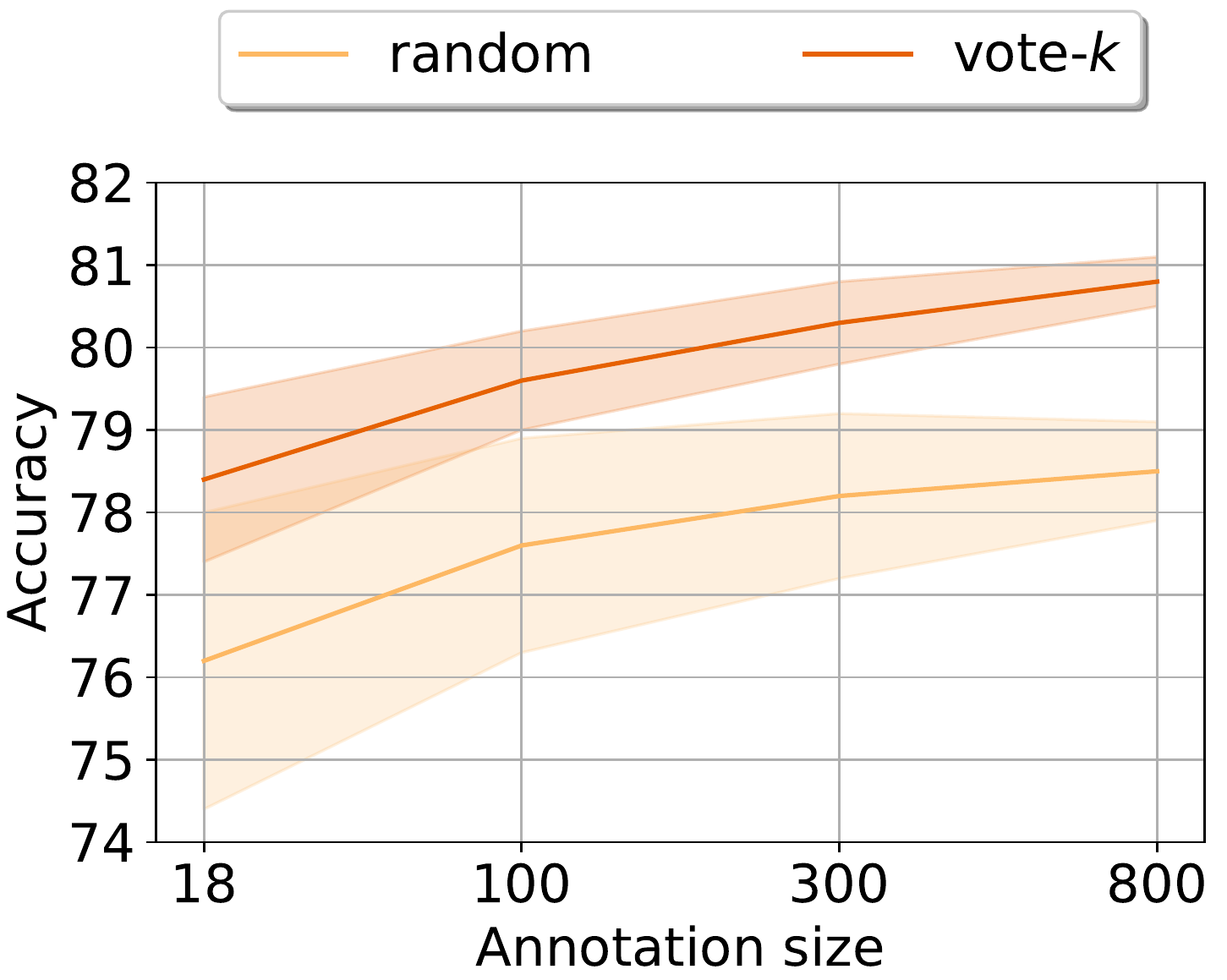}
\caption{OPT-175B performance of ICL-random and ICL-\votek on HellaSwag 
}
\label{fig:hellaswag_opt}
\end{figure}

\section{Removing outliers for finetuning}

Here we show that explicitly removing outliers also helps finetuning to benefit from vote-$k$. 

\label{app:remove_outlier_FT}
\begin{table}[h!]
\addtolength{\tabcolsep}{-4.0pt}
\centering
\begin{tabular}{@{}ccccc@{}}
\toprule[.1em]
& \multicolumn{2}{c}{Outliers not removed } & \multicolumn{2}{c}{10\% outliers removed } \\
\cmidrule(r){2-3}\cmidrule(l){4-5}
& FT-random
& FT-\votek
& FT-random
& FT-\votek
\\

\midrule[.1em]

HellaSwag
& 55.6%\tiny{3.4}
& 53.5%\tiny{1.6}
& 56.8%\tiny{2.6}
& 59.6%\tiny{1.3}
\\

MRPC
& 56.3%\tiny{6.8}
& 55.6%\tiny{2.8}
& 57.9%\tiny{4.2}
& 60.4%\tiny{2.0}
\\

\bottomrule[.1em]
\end{tabular}
\caption{Effects of \votek in finetuning(FT) with annotation budget of 100. \textit{10\% outliers removed} means that we removed 10\% of examples farthest to the training data, measured by average cosine similarity. The selection is conducted after example removal. After removing outliers, \votek selection improves the model few-shot performance.
}
\label{tab:outlier_FT}
\end{table}

\section{Diversity and Representativeness of Selected Samples}
\label{sec:div_rep_analysis}
We hypothesized that both representativeness and diversity are crucial for \firststep (\S\ref{sec:sample_selection_method}).
Here we evaluate the diversity and representativeness of samples that are selected by different methods, using the methods from prior work on active learning \citep{margatina2021active}; their measures of diversity and representativeness use token overlap or embedding cosine similarities. 
As shown in Table \ref{tab:diverse_represent}, \votek improves both the diversity and the representativeness as compared to random selection. %achieves a balanced middle ground between the two properties. 
%The embedding diversity method yields more diversity at the expense of reduced representativenss, resulting in downstream perform drops. 
%\jungo{Hongjin's TODO: add self-dissimilar (embedding div.)?}
%% \jungo{maybe drop. this section entirely??}

\begin{table}[h!]
\addtolength{\tabcolsep}{-4.0pt}
\centering
\small
\begin{tabular}{@{} c   m{0.001em}  ccc m{0.001em}  ccc m{0.001em}  ccc }
\toprule[.1em]

 \textbf{Method}
&& 
 \multicolumn{3}{c}{\textbf{DIV-I}}
&& 
 \multicolumn{3}{c}{\textbf{DIV-F}}
&& 
 \multicolumn{3}{c}{\textbf{REPR.}}
\\
\cmidrule(lr){1-1}
\cmidrule(lr){3-5}
\cmidrule(lr){7-9}
\cmidrule(lr){11-13}

Selection
&&  HellaSwag
& SST-5 
& MWoZ
&& HellaSwag
& SST-5
& MWoZ
&& HellaSwag
& SST-5
& MWoZ
\\

\midrule[.1em]

Random
&& 0.182\tiny{0.007}
& 0.099\tiny{0.003}
& 0.368\tiny{0.008}
&& 0.415\tiny{0.008}
& 0.317\tiny{0.004}
& 0.675\tiny{0.006}
&& 0.558\tiny{0.007}
& 0.424\tiny{0.003}
& 0.696\tiny{0.004}
\\

\Votek
&& 0.191
& 0.108
& 0.379
&& 0.425
& 0.321
& 0.683
&& 0.565
& 0.426
& 0.702
\\

\bottomrule[.1em]
\end{tabular}
\caption{DIV-I refers to diversity in input space, which measures the diversity of selected data in the input feature space, i.e., raw text; DIV-F refers to diversity in feature space, which measures the diversity in the dense feature space, i.e., sentence embeddings; REPR. refers to representativeness, which measures the representativeness of selected data. Subscripts stand for standard deviation. 
}
\label{tab:diverse_represent}
\end{table}

\section{Details of \FirstStep Methods}
\label{sec:details-sample-selection}
In this section, we provide details of \firststep methods used in Section~\ref{subsec:sample-selection-methods}. 
% \rui{Why do we have discuss this section?}
% \chen{Mari suggested this baseline, and we are experimenting with it. }

\subsection{\Votek \FirstStep}
Algorithm~\ref{alg:vote-k} describes the \votek \firststep method introduced in Section~\ref{sec:sample_selection}. 

\subsection{Greedy Algorithm for Maximizing Facility Location}
\label{sec:Submodularity-Based Sample Selection}
\citet{Lin2009HowTS} proposed to maximize the facility location objective to optimize representativeness of the selected samples. Since this objective satisfies the submodular property, they applied a greedy algorithm as an approximation. 
Algorithm~\ref{alg:submodularity} describes the \firststep method adapted from this greedy algorithm. 
%\chen{check the claim of diversity. Diversity seems to be at most implicitly encouraged... } 

\subsection{Embedding Diversity}
\label{sec:Embedding Diversity}
%\jungo{renamed self dissimilar to embedding diversity for better readability. Maybe change later.}
This method aims to find diverse samples to annotate using embedding vectors.
We first compute a vector representation for each \emph{unlabeled} training instance by Sentence-BERT \citep{reimers-gurevych-2019-sentence}, which is a variant of BERT \citep{devlins2019bert}, finetuned to detect paraphrases.\footnote{\url{https://huggingface.co/sentence-transformers/all-mpnet-base-v2}.}
For instance, consider an example from SST-5 sentiment analysis in Table \ref{tab:datasets_all}:\textit{A very well-made, funny and entertaining picture}.
We simply run Sentence-BERT on this text input and average the resulting vectors over the words to obtain a vector representation.

Once embeddings are computed for all training data, we use them to find a diverse set of training instances. 
The intuition here is that a diverse set of annotated examples facilitates the subsequent prompt retrieval step since similar in-context examples can be found for many test instances.
To find a set of diverse embeddings, we take a simple, iterative approach: in every iteration, we choose an instance furthest from the already chosen ones.
Specifically, let $\mathcal{L}$ and $\mathcal{U}$ denote the sets of already chosen (i.e., labeled) samples and unlabeled samples, respectively.
Suppose also that $M$ is the target number of labeled examples (i.e., the annotation budget).
Then, in every iteration, we choose the unlabeled sample that has the largest total cosine distance from $\mathcal{L}$: $\argmin_{u \in \mathcal{U}}\sum_{\ell \in \mathcal{L}} cos(u, \ell)$.
Here we abuse $u$ and $\ell$ to mean both the instances and their embedding vectors from Sentence-BERT.
The first labeled sample is randomly selected from the 3K unlabeled examples (\S\ref{sec:datasets_tasks}), and the iterative process continues until $|\mathcal{L}|\!=\!M$.

\begin{algorithm}[t]
\small
\caption{Voke-k \FirstStep}
\label{alg:vote-k}
\begin{algorithmic}[1]
\State \textbf{Input:} $\mathcal{X} = \{x_i\}_{i=1}^{N}$: a set of unlabeled samples; $M$: the number of samples to be selected; LM: inference language model. 
\State \textbf{Initialization:} $\mathcal{L} = \varnothing$, $\mathcal{U} = \mathcal{X}$. $G=(V, E)$, where $V = \mathcal{X}$ and $(u, v) \in E$ if $v$ is one of $u$'s $k$ nearest vertices in terms of the cosine similarity between the embeddings. %\nascomment{don't you need to construct the graph, too?}
\While{$|\mathcal{L}| < M / 10$}
    \State $u^{*}=\arg \max_{u \in \mathcal{U}} \sum_{v \in \{v | (v, u) \in E, v \in \mathcal{U}\}} s (v), \quad \text{where} \ s(v) = \rho ^{- |\{\ell \in \mathcal{L}| (v, \ell) \in E \}|}, \quad \rho > 1$
    \State $\mathcal{L} = \mathcal{L} \cup\left\{u^{*}\right\}$
    \State $\mathcal{U} = \mathcal{U} \setminus\left\{u^{*}\right\}$
\EndWhile \label{line:first}
% \For{$u$ in $\mathcal{U}$}
%     \State $\mathrm{score}(u)$ = LM$(\mathcal{L}, u)$ 
%     % \nascomment{see comments in main text; this is too vague}
% \EndFor
% \State $\mathrm{indices}$ = $\arg\mathrm{sort}_{u \in \mathcal{U}}$ $\mathrm{score}(u)$
% \For{$u$ in $\mathcal{U}$}
%     \State $\mathrm{score}(u)$ = $\left\{\begin{array}{ll}LM(\mathcal{L}, u) \\ LM(\mathcal{L}, u) \end{array}\right.$
%     % \nascomment{see comments in main text; this is too vague}
% \EndFor
\For{$u$ in $\mathcal{U}$} \label{line:lm-start}
    \State $\mathrm{score}(u)$ = $\frac{1}{\textbf{q}}\sum_{t} \textrm{log} p(q_t|\textbf{q}_{<t},\textbf{z};\Theta) $, where $p$ is LM prediction function and $\Theta$ is LM parameters
\EndFor \label{line:lm-ends}

% \sigma(s,i) = \left\{\begin{array}{ll}\tau_{si} & \mbox{si } \{s,i\} \in E \\ \infty & \mbox{sinon.} \end{array}\right.

\For{$j = 1, \ldots, 9$} \label{line:it-start}
    % \State $\mathcal{U}_j$ is ... \chen{check with Hongjin: the precise definition of the bucket $\mathcal{U}_j$. }
    \State $\mathcal{U}_j = \mathrm{indices}[(j-1)|\mathcal{U}|/10:j|\mathcal{U}|/10]$ 
    \For{$i = 1, \ldots, |\mathcal{U}_j|$}
        \State $u^{*}=\arg \max_{u \in \mathcal{U}_j} \sum_{v \in \{v | (v, u) \in E, v \in \mathcal{U}_j\}} s (v), \quad \text{where} \ s(v) = \rho ^{- |\{\ell \in \mathcal{L}| (v, \ell) \in E \}|}, \quad \rho > 1$ 
        % \chen{check with Hongjin: in each percentile, is the unlabeled set updated? } 
        \State $\mathcal{L} = \mathcal{L} \cup\left\{u^{*}\right\}$
        \State $\mathcal{U}_j = \mathcal{U}_j \setminus\left\{u^{*}\right\}$
    \EndFor
\EndFor\label{line:it-ends}
\State \textbf{Return:} $\mathcal{L}$: selected samples.
\end{algorithmic}
\end{algorithm}

\begin{algorithm}[t]
\small
\caption{Greedy Algorithm for Facility Location Objective}
\label{alg:submodularity}
\begin{algorithmic}[1]
\State \textbf{Input:} $\mathcal{U} = \{x_i\}_{i=1}^{N}$: a set of unlabeled samples; $M$: the number of samples to be selected. 
\State \textbf{Initialization:} $\mathcal{L} = \varnothing$, $\mathcal{U} = V$. $\forall i, \rho_i = -1$: maximum similarity of $x_i$ to selected samples. 
\While{$|\mathcal{L}| < M$}
    \State $u^{*}=\arg \max_{u \in \mathcal{U}} \sum_{i = 1}^{N}\left(\max \left\{0, \cos(x_i, x_u) - \rho_{i}\right\}\right)$
    \State $\mathcal{L} = \mathcal{L} \cup\left\{u^{*}\right\}$
    \State $\mathcal{U} = \mathcal{U} \setminus\left\{u^{*}\right\}$
    \State $\forall i, \rho_{i}=\max \left\{\rho_{i}, \cos(x_i, x_{u^*})\right\}$ \quad // update maximum similarity of each $x_i$ to selected samples
\EndWhile
\State \textbf{Return:} $\mathcal{L}$: selected samples.
\end{algorithmic}
\end{algorithm}

\end{appendices}

\end{document}